%% file: main.tex
\documentclass[10pt,twocolumn,letterpaper]{article}
\usepackage{amssymb}%
\usepackage{pifont}%
\newcommand{\cmark}{\ding{51}}%
\newcommand{\xmark}{\ding{55}}%

\usepackage{graphicx}
\usepackage{booktabs}
\usepackage{threeparttable}
\usepackage[accsupp]{axessibility}

\usepackage{multirow} 
\usepackage{makecell} 
\usepackage{caption}
\usepackage{wrapfig}
\usepackage{lipsum}
\usepackage{diagbox}
\usepackage{colortbl}
\usepackage[pagenumbers]{cvpr}
\usepackage{amsmath}
\usepackage{dsfont}
\DeclareMathOperator*{\argmax}{argmax}
\DeclareMathOperator*{\argsort}{argsort}
\usepackage{color}
\usepackage{float}
\usepackage{threeparttable}
\input{preamble}

\definecolor{cvprblue}{rgb}{0.21,0.49,0.74}
\usepackage[pagebackref,breaklinks,colorlinks,citecolor=cvprblue]{hyperref}

\newcommand{\figref}[1]{Fig.~\ref{#1}}
\newcommand{\tabref}[1]{Table~\ref{#1}}
\newcommand{\equref}[1]{Eq.~(\ref{#1})}
\newcommand{\secref}[1]{Sec.~\ref{#1}}

\def\paperID{15} %
\def\confName{CVPR}
\def\confYear{2024}

\title{Rethinking Open-World Semi-Supervised Learning: \\ Distribution Mismatch and Inductive Inference}

\author{Seongheon Park$^*$ \quad\quad Hyuk Kwon$^*$ \quad\quad Kwanghoon Sohn \quad\quad Kibok Lee \\ 
Yonsei University \\
{\tt\small {\{sam121796,kh12043,khsohn,kibok\}@yonsei.ac.kr}}}

\begin{document}
\maketitle
\def\thefootnote{*}\footnotetext{Authors contributed equally to this work.}
\input{sec_workshop_camera/0_abstract}

\input{sec_workshop_camera/1_intro}
\input{sec_workshop_camera/1_1_ROWSSL}

\input{sec_workshop_camera/2_methods_ver2}
\input{sec_workshop_camera/3_experiments}

\input{sec_workshop_camera/workshop_suppl_ver2}
{
    \small
\bibliographystyle{ieeenat_fullname}
\bibliography{CVPRW_arXiv/main}
}

\end{document}

%% file: preamble.tex
\usepackage[dvipsnames]{xcolor}
\usepackage{multirow}

%% file: sec_workshop_camera/0_abstract.tex
\begin{abstract}
\vspace{-8pt}
Open-world semi-supervised learning (OWSSL) extends conventional semi-supervised learning to open-world scenarios by taking account of novel categories in unlabeled datasets.
Despite the recent advancements in OWSSL, the success often relies on the assumptions that
1) labeled and unlabeled datasets share the same balanced class prior distribution, which does not generally hold in real-world applications, and
2) unlabeled training datasets are utilized for evaluation, where such transductive inference might not adequately address challenges in the wild.
In this paper, we aim to generalize OWSSL by addressing them.
Our work suggests that practical OWSSL may require different training settings, evaluation methods, and learning strategies compared to those prevalent in the existing literature.
\end{abstract}
\vspace{-12pt}

%% file: sec_workshop_camera/1_intro.tex
\section{Introduction}
\label{sec:intro}

OWSSL has been introduced to discover novel classes within an unlabeled dataset while accurately classifying known classes.
However, we argue that OWSSL may not reflect real-world scenarios for the following reasons:
1) recent works on OWSSL assume balanced and identical class prior distribution between labeled and unlabeled datasets during the learning process, and
2) they only consider a transductive learning setting, which focuses on categorizing instances from the unlabeled training datasets.

\begin{figure}[t]

\subfloat[
In ROWSSL, we consider the cases when the class prior of labeled and unlabeled datasets are matched (\textbf{left}) and mismatched (\textbf{right}).
]{
	\label{fig:1_a}
\includegraphics[width=0.45\textwidth]{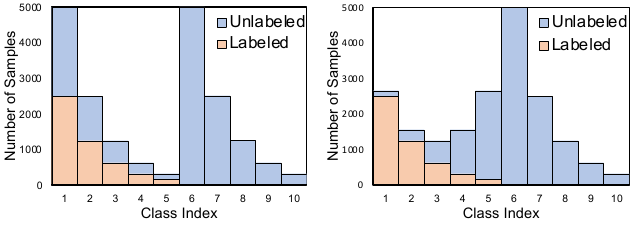} }
\hfill
\subfloat[Examples of transductive and inductive inference in ROWSSL.
Inductive inference is performed without looking at other test data.
]{
	\label{fig:1_b}
 
 \includegraphics[width=0.45\textwidth]{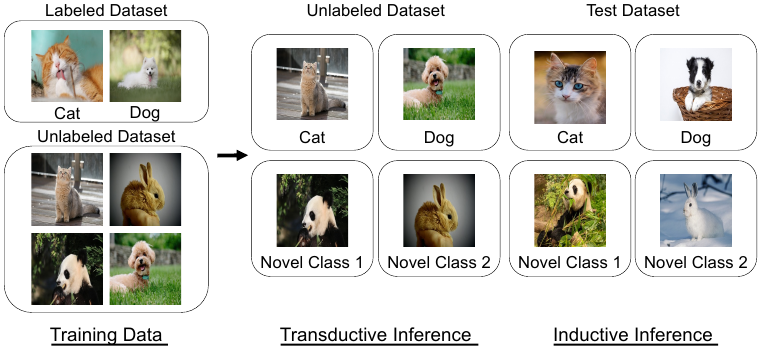} } 
 \vspace{-5pt}
\caption{Examples of scenarios considered in ROWSSL.}
\label{fig:1}
\vspace{-20pt}
\end{figure}

Indeed, in-the-wild data naturally follow a long-tailed distribution and are exposed to label distribution shifts~\cite{oliver2018realistic,hu2022non}, \ie, labels are missing not at random (MNAR; \figref{fig:1_a} right) rather than missing completely at random (MCAR; \figref{fig:1_a} left).
Class prior distribution mismatch between labeled and unlabeled datasets happens for multiple reasons, \eg, the data distribution itself could change over time, or annotators might prefer to annotate relatively easy classes or they could miss difficult classes.
However, most OWSSL methods assume a balanced class prior for training, which often hampers performance when the assumption does not hold.
Also, most OWSSL methods assume a transductive learning setting, which is specialization on given unlabeled training data rather than generalization on unseen test data as illustrated in \figref{fig:1_b}.
While transductive learning is useful for category discovery,
it does not guarantee reliable performance when classifying discovered categories from unseen test data.
Instead, inductive learning is important in safety-critical applications such as medical diagnosis, \eg, a model that can discover novel diseases in a specific patient cohort might still misclassify diseases in unseen patients.

To this end, we extend OWSSL by addressing such practical training and evaluation settings, coined \textbf{R}ealistic \textbf{O}pen-\textbf{W}orld \textbf{S}emi-\textbf{S}upervised \textbf{L}earning (ROWSSL).
In this task, we consider long-tailed distribution with class prior distribution mismatch between labeled and unlabeled datasets for training, and inductive and transductive inferences for evaluation.
To address the aforementioned challenges, we introduce \textbf{D}ensity-based \textbf{T}emperature scaling and \textbf{S}oft pseudo-labeling (DTS)
to learn class-balanced representations taking account of local densities and
reduce classifier bias toward the head and known classes simultaneously.
To achieve this, we propose to measure the tailedness as a proxy for the unknown class prior via density estimation on the representation space.
With these proxies, 
we introduce a dynamic temperature scaling approach for balanced contrastive learning, which dynamically adjusts the temperature parameter 
for each anchor to shape a representation space to have better linear separability between head and tail classes.
Also, we address the classifier bias via class uncertainty-aware soft pseudo-labeling, considering a density variance as an uncertainty measure.

%% file: sec_workshop_camera/1_1_ROWSSL.tex
\section{ROWSSL}
\label{sec:GCDW}

\subsection{Training setting}
Suppose we have a partially labeled dataset $\mathcal{D} = \mathcal{D}_l \cup \mathcal{D}_u$,
where $\mathcal{D}_l = \{\mathbf{x}_i,\mathbf{y}_i\}_{i=1}^{N_l}\in \mathcal{X} \times \mathcal{Y}_l$ is the labeled dataset with $N_l$ samples which belongs to one of the $C_{\text{old}}$ known classes, and
$\mathcal{D}_u = \{\mathbf{x}_i\}_{i=1}^{N_u}\in \mathcal{X}$ is the unlabeled dataset with $N_u$ samples,  with its underlying label space $\mathcal{Y}_u$ containing both of $C_{\text{old}}$ known classes and $C_{\text{new}}$ novel classes.
In GCDW, the labeled dataset $\mathcal{D}_l$ has a long-tailed distribution with an imbalance ratio $\gamma_l>1$, while the unlabeled dataset can have an arbitrary class prior, including MCAR and MNAR scenarios as depicted in~\figref{fig:1_a}.
Following \cite{vaze2022generalized,wen2022simple}, the total number of classes $C = C_{\text{old}} + C_{\text{new}}$ is either assumed to be known a priori
or estimated through off-the-shelf methods~\cite{han2019learning}.
Our objective is to train a parametric classifier $f:\mathbb{R}^d \mapsto [0,1]^{C}$ to correctly assign class labels to both known and novel classes.

\subsection{Evaluation setting and metrics}
In \tabref{tab:bacon_eval}, we compare the balanced overall accuracy of previous methods with different evaluation strategies on CIFAR-100-LT, where $\ddagger$ indicates the result aligned with \cite{bai2023towards}, with a maximum discrepancy of $1.3\%$.
The ratio of the number of known and novel classes is 80:20 for Split~1 and 50:50 for Split~2.
\paragraph{Transductive inference.}
Prior works~\cite{vaze2022generalized,cao2021open} have performed transductive inference for their methods on the unlabeled training dataset (``Train'' evaluation set in \tabref{tab:bacon_eval}).
Following them, we measure the clustering accuracy between the ground truth labels $y_i$ and the model's predictions $\hat{y_i}$ through the Hungarian algorithm~\cite{kuhn1955hungarian}:
\begin{equation}\label{equ:acc}
\text{ACC} = \frac{1}{|\mathcal{D}_u|}\sum_{i=1}^{|\mathcal{D}_u|} \mathds{1}\{y_i=p^*(\hat{y_i})\},
\end{equation}
where $p^*$ is the optimal permutation that matches the predicted cluster assignments to the ground truth class labels.
While transductive learning is useful for category discovery, 
it does not guarantee the reliable performance of the learned model to classify discovered categories in unseen test data.

\paragraph{Transductive inference on the test set.}

BaCon~\cite{bai2023towards} utilizes a balanced test dataset following the common practice in long-tailed recognition.
However, they perform $k$-means clustering on the entire test dataset for evaluation, \ie, the classification result depends on other test data, which corresponds to transductive inference (``Recluster \checkmark~and Rematch \checkmark'' in \tabref{tab:bacon_eval}).
Also, they ignore the clusters found during training and match the clusters of the test set with the classes, resulting in unintentional concept shifts, \eg, the cat class during training might be matched with the lion class at test time.
Hence, their evaluation results do not properly reflect the generalizability of the models to online inference, which is often required in real-world scenarios, and they cannot identify the semantics of classes.
Furthermore, $k$-means assumes the presence of the uniform cluster prior, which leads to biased results in relation to the balanced test set statistics~\cite{assran2022hidden}.
We found that the high performance of BaCon might be due to the uniform prior assumption of $k$-means and concept shifts by rematching for the best performance, \eg, when evaluated on the imbalanced training dataset, BaCon is on par with other methods in Split~1, and outperformed by other methods in Split~2.

\begin{table}
    \centering
    \caption{Comparison of different evaluation strategies.
    \vspace{-5pt}
    }
     \resizebox{0.75\columnwidth}{!}
    {%
    \begin{tabular}{ccccccccc}
        \toprule
        Data split & \multicolumn{4}{c}{Split~1 \cite{bai2023towards}}  & \multicolumn{4}{c}{Split~2} \\
        \cmidrule(lr){1-1} \cmidrule(lr){2-5} \cmidrule(lr){6-9}
         Eval set & Train & \multicolumn{3}{c}{Test} & Train & \multicolumn{3}{c}{Test} \\
         \cmidrule(lr){1-1} \cmidrule(lr){2-2} \cmidrule(lr){3-5} \cmidrule(lr){6-6} \cmidrule(lr){7-9}
         Recluster & - & \checkmark & \xmark & \xmark & - & \checkmark & \xmark & \xmark \\
         \cmidrule(lr){1-1} \cmidrule(lr){2-5} \cmidrule(lr){6-9}
          Rematch & - & \checkmark & \checkmark & \xmark & - & \checkmark & \checkmark & \xmark\\
         \midrule
         $k$-means &  37.8 & 55.0& 38.7& 37.7 &  34.2& \underline{55.0}& 37.0&30.3\\
         ORCA &  38.9& 51.2${}^{\ddagger}$ &49.1 & 42.4 & 25.0& 34.9& 32.0&29.5\\
         GCD &  49.5& \underline{63.5}${}^{\ddagger}$ & 50.3 & \underline{48.5} &  42.3& 54.4& 40.5&\underline{38.1}\\
         SimGCD &  49.2& 61.3 & \underline{52.5}${}^{\ddagger}$ & 44.9 & \underline{46.5}& 50.2&42.9&37.4\\
         BaCon & \underline{50.8}& \textbf{67.9}${}^{\ddagger}$ & 50.7 &47.5 &  38.0& \textbf{59.2}& \underline{45.2}&35.9\\
         \midrule
        \rowcolor{cyan!10} Ours &  \textbf{54.1}& 61.7& \textbf{55.8} & \textbf{52.1} & \textbf{53.7} &  51.9 & \textbf{53.7}& \textbf{48.1}\\
         \bottomrule
    \end{tabular}
    }
    \label{tab:bacon_eval}
\vspace{-20pt}
\end{table}

\paragraph{Inductive inference.}
To evaluate the generalizability of models, we consider inductive inference. %
Specifically, we evaluate the models on the disjoint test dataset by nearest centroid classification, where the center of clusters found by optimal matching $p^*$ from~\eqref{equ:acc} on the training set are utilized as parametric class centers (``Recluster \xmark~and Rematch \xmark'' in \tabref{tab:bacon_eval}).
To confirm that concept shifts are beneficial to maximize the performance, we also apply Hungarian matching between the parametric clustering results with the classes (``Recluster \xmark~and Rematch \checkmark'' in \tabref{tab:bacon_eval}).
While rematching results in better performance, this ignores the semantics of categories discovered during training.
In fact, rematching corresponds to transductive inference, as it requires gathering the parametric clustering results.
Throughout experiments, we focus on evaluation without reclustering and rematching for inductive inference.

%% file: sec_workshop_camera/2_methods_ver2.tex
\section{Proposed Method}
\label{sec:method}

We propose an end-to-end approach that jointly learns the representation and parametric classifier, similar to Wen~\etal \cite{wen2022simple}.
The network architecture is composed of an encoder $E$ followed by two heads $f$ and $g$.
The encoder $E$ can be a pre-trained model, \eg, a ViT pre-trained with DINO \cite{caron2021emerging},
$\mathbf{z} = E(\mathbf{x}) \in \mathbb{R}^d$
is a feature vector representing the input image $\mathbf{x}$,
$f$ is an $\ell_2$-normalized linear classifier, and
$g$ is a multi-layer perceptron (MLP) projecting $\mathbf{z}$ to a lower dimensional vector $\mathbf{h}$ for representation learning.

\subsection{Training objectives} \label{sec:3.2}
\paragraph{Representation learning.}
We adopt contrastive learning (CL) loss for representation learning.
From a mini-batch $B$, two views of an image are obtained through random augmentation, represented as $\mathbf{x}$, and $\mathbf{x}'$.
These images are then fed into the query and key networks
$E \circ g$ and $E' \circ g'$, yielding a pair of $\ell_2$-normalized embeddings $\mathbf{h} = (E \circ g)(\mathbf{x})$ and $\mathbf{b} = (E' \circ g')(\mathbf{x}')$, respectively, 
where the key network is updated by exponential moving average (EMA), following MoCo~\cite{he2020momentum}.
Self-supervised learning loss is defined as:
\begin{equation}
\label{eq:infonce}
l_{u}(\mathbf{x}_i)=-\log{\frac{\exp{(\mathbf{h}_i\cdot \mathbf{b}_{+}/\tau})}{\sum_{\mathbf{b}'\in \mathbf{Q}}\exp{(\mathbf{h}_i\cdot \mathbf{b}'/\tau})}}.
\end{equation}
Here, $\mathbf{b}_{+}$ is a positive key, and 
the queue $\mathbf{Q}=\{\mathbf{b}_j\}_{j=1}^{Q}$ is updated sequentially with key embeddings $\mathbf{b}$ following the first-in-first-out (FIFO) scheme, where $Q$ is the predefined queue size.
$\mathbf{Q}=\{\mathbf{b}_j\}_{j=1}^{Q}$ is a queue that contains the key embeddings $\mathbf{b}$ of a predefined size Q.
For effective utilization of label information, we adopt the variation of the supervised contrastive loss 
$l_{\text{sup}}(\mathbf{x}_i,\mathbf{y}_i)$ 
\cite{khosla2020supervised} which maintains multiple positive pairs on-the-fly by comparing the query label to a label queue \cite{9945500}.
Overall representation learning loss is defined as:
\begin{equation}
L_{\text{rep}}=(1-\lambda_{\text{rep}})\frac{1}{|B|}\sum_{i\in B}l_{u}(\mathbf{x}_i)+\lambda_{\text{rep}}\frac{1}{|B_l|}\sum_{i\in B_l}l_{\text{sup}}(\mathbf{x}_i,\mathbf{y}_i),
\end{equation}
where $B_l$ corresponds to the labeled subset of $B$ and $\lambda_{\text{rep}}$ is a balancing factor.

\paragraph{Classifier learning.}
Our parametric classification framework follows the self-distillation methods \cite{caron2021emerging}.
We employ a prototypical classifier where the weight parameters of linear classifier $f$ are regarded as cluster centroids.
To discover novel classes and allocate each sample to the optimal cluster, we condition the cluster centroids to contain class information through multi-tasking self-supervised and supervised objectives \cite{fini2023semi}.
Classification loss is defined as:
\begin{equation}
\label{eq:classifier}
l_{\text{cls}}(\mathbf{x}_i,\mathbf{y}_i)=-\sum_{k=1}^{C}\mathbf{\bar{y}}_i^k\log{(\mathbf{p}_i^k)},\quad \mathbf{\bar{y}}_i=  \begin{cases}
    \mathbf{y}_i, & \text{$\mathbf{x}_i\in \mathcal{D}_l$},\\
    \mathbf{q}_{i}, & \text{$\mathbf{x}_i\in \mathcal{D}_u$},
  \end{cases}
\end{equation}
where $\mathbf{p}=\text{softmax}(f(\mathbf{z})/\tau_s)$ is the temperature-scaled softmax probability with $\tau_s$,
$\mathbf{y}$ is a one-hot representation of the ground-truth label, and the soft pseudo-label $\mathbf{q}=\text{softmax}(f(\text{sg}(\mathbf{z'}))/\tau_t)$ is produced by another augmented view of $\mathbf{x}$ through sharpening, \ie, $\tau_s>\tau_t$.
Following \cite{wen2022simple}, we also adopt a mean-entropy maximization regularizer $H(\mathbf{\Bar{p}})=\sum_{k=1}^{C}\mathbf{\bar{p}}^k\log{(\mathbf{\Bar{p}}^k)}$, where $\mathbf{\Bar{p}}=\frac{1}{2|B|}\sum_{i\in B}(\mathbf{p}_i+\mathbf{p}'_i)$,
to avoid an inactivation of classifier heads.
The classifier learning loss is defined as:
\begin{equation}
L_{\text{cls}} = \frac{1}{|B|}\sum_{i\in B}l_{\text{cls}} (\mathbf{x}_i,\mathbf{y}_i) -\varepsilon H(\mathbf{\Bar{p}}),
\end{equation}
where $\varepsilon$ controls the weight of the regularizer.
Overall training objective is %
defined as:
$L_{\text{rep}}+L_{\text{cls}}$.

\subsection{Constructing tailedness prototypes} \label{sec:3.3}
\paragraph{Tailedness estimation.}
Different from prior OWSSL settings, the true class prior is unknown in ROWSSL, as MNAR is considered.
To learn a model without knowing the true class prior, 
we define ``tailedness'' as a surrogate for the class prior based on density estimation within the representation space. 
Since tail classes often exhibit lower intra-class consistency than head classes, samples of tail classes tend to sparsely distribute on the representation space~\cite{bai2023effectiveness,liu2022selfsupervised}.
Building on this, we learn tailedness prototypes, aiming to explore stable and efficient proxies to discover tail class samples.
To begin with, we utilize the queue $\mathbf{Q}=\{\mathbf{b}_i\}_{i=1}^{Q}$ of the CL branch in \secref{sec:3.2}, as the neighbors in the entire dataset cannot be captured by looking at only a mini-batch.
We initialize $\ell_2$-normalized tailedness prototypes $\mathbf{M}=\{\mathbf{m}_i\}_{i=1}^{M}$ by $k$-means on the features of the queue, and estimate density $d_i$ of a prototype $\mathbf{m}_i$ based on the weighted average of the cosine similarity of its $K$-nearest neighbors:
\begin{equation} \label{eq:density}
d_j^K = \frac{1}{\sum_{k=1}^{K}w_k}\sum_{i\in \mathcal{N}_K(\mathbf{m}_j)}
w_i (\mathbf{m}_j \cdot \mathbf{b}_i),
\end{equation}
where $\mathcal{N}_K(\mathbf{m}_j)$ is the set of the indices of the $K$-nearest neighbors of $\mathbf{m}_j$, 
and the distance-based weighting
$w_i = \argsort_j (\mathbf{m}_j \cdot \mathbf{b}_i)$
to reflect the local density better, reducing the effect of noisy density estimation \cite{Dudani1976wknn}.
Tailedness score $s_i$ of each sample $\mathbf{x}_i$ is defined as:
\begin{equation}
 s_i=d_{j^*(i)}, \quad \text{where } j^*(i) = \argmax_j \ \mathbf{m}_j \cdot \mathbf{b}_i.
\end{equation}

\paragraph{Prototype update.}
We update tailedness prototypes by EMA for stable learning.
Specifically, the queue $\mathbf{Q}$ is split into a disjoint set of key features $\{\mathbf{U}_j\}_{j=1}^M$, where each key feature is assigned to the nearest tailedness prototype:
  \begin{equation}\label{equ:update}
\mathbf{m}_j \leftarrow \text{normalize}\left[\lambda_{\text{tail}}\mathbf{m}_j + (1-\lambda_{\text{tail}})\frac{1}{|\mathbf{U}_j|}\sum_{\mathbf{b}'\in{\mathbf{U}_j}}\mathbf{b}'\right],
 \end{equation}
where $\mathbf{m}_j$ is $\ell_2$-normalized and $\lambda_{\text{tail}}$ is a momentum coefficient.

\subsection{Density-based learning strategy} \label{sec:3.4}

\paragraph{Dynamic temperature scaling.}
We aim to handle long-tailed data through self-supervised representation learning by controlling temperature parameter $\tau$, which has been shown to play a significant role in learning good representations~\cite{wang2020understanding}.
Specifically, we view the contrastive loss through the average distance maximization perspective~\cite{kukleva2023temperature}.
From this view, a large $\tau$ allows the model to maximize the average distance across a wide range of neighbors, which is advantageous for preserving local semantic structures.
On the other hand, a small $\tau$ helps to learn instance-specific features by encouraging a uniform distribution of embeddings across the representation space.
Based on this perspective, we present a novel representation learning method, the dynamic temperature scaling for CL.
Specifically,
we adjust the temperature parameter $\tau$ in \eqref{eq:infonce} as a function of the anchor's tailedness score $s_i$:
\begin{equation}\label{equ:temp}
\tau(\mathbf{x}_i)=\tau_{\text{min}}+\frac{s_i-\text{min}_t(d_{t})}{\text{max}_t(d_{t})-\text{min}_t(d_{t})}(\tau_{\text{max}}-\tau_{\text{min}}),
\end{equation}
where $\tau_{\text{min}}$ and $\tau_{\text{max}}$ are hyperparameters, denoting the minimum and maximum values of temperature, respectively.
As tail classes benefit from learning instance-specific features while head classes are required to preserve their local semantic structure~\cite{kukleva2023temperature}, our approach dynamically assigns smaller $\tau$ to tail classes and larger values to head classes.
This allows the model to learn class-balanced representations, achieving better linear separability between the long-tailed classes without knowing the true class prior.
\vspace{-5pt}
\paragraph{Class uncertainty-aware soft pseudo-labeling.}
For pseudo-labeling in long-tailed recognition, the distribution of pseudo-labels on unlabeled data tends to be biased toward head classes~\cite{arazo2020pseudo}.
For conventional long-tailed recognition, the effect of bias can be mitigated by giving more weight to tail classes inversely proportional to their class prior~\cite{menon2020long}.
However, this approach might not work well in ROWSSL, as pseudo-labels tend to be biased toward known classes, such that they are often more biased toward known-tail classes than novel-head classes~\cite{wang2022partial}.
To this end, we propose to adjust the soft pseudo-label $\mathbf{q}_i$ in \eqref{eq:classifier} with regard to the class uncertainty.
Intuitively, for classes that are easy to learn, their samples will consistently be assigned to a specific tailedness prototype.
Conversely, samples from more difficult and uncertain classes will be arbitrarily distributed across various prototypes.
Based on this idea, we propose to use the standard deviation of tailedness scores among samples within each class as a measure of the relative learning uncertainty of each class as the additive class uncertainty.
At each training iteration, we gather the tailedness scores in the dataset with respect to each sample's pseudo-label into the class-wise tailedness queue $S^c$.
We define the class uncertainty vector of the $e$-th training iteration $\mathbf{u}_e=[u^1, \dots, u^C]$, $e=1,..., E$, as a collection of the standard deviation of tailedness scores per class:
\begin{equation}
u^c=\text{std}(S^c) \text{ where }
S^c = \{s_i \mid x_i \in \mathcal{D}, \,\argmax_k(\mathbf{\bar{y}}_i^k) = c\}.
\end{equation}
Note that $u^c = 0$ when $S^c = \varnothing$ and $\mathbf{u}_0 = \mathbf{0}$.
Then, we adjust the output of the classifier with the class uncertainty:
\begin{equation}
\mathbf{q}_i=\text{softmax}[(f(\text{sg}(\mathbf{z}_{i}')+\lambda_{\text{var}}\mathbf{u}_{e-1})/\tau_t],
\end{equation}
where $\lambda_{\text{var}}$ is a hyperparameter.
Our approach can be considered as a variation of the uncertainty-adaptive margin loss in~\cite{cao2021open}, 
which mitigates classifier bias towards the head and known classes in a unified way.

\vspace{-5pt}

%% file: sec_workshop_camera/3_experiments.tex
\section{Experimental Results} 
\label{sec:experiments}
\begin{table}[t]
    \centering
        \caption{Results on CIFAR-100-LT. \textbf{Tr}: transductive, \textbf{In}: inductive,
        \textbf{ACC}: average accuracy in \equref{equ:acc},
        \textbf{bACC}: average of per-class accuracy. %
        The best and second-best results are highlighted in \textbf{bold} and \underline{underlined}, respectively.
        }
            \vspace{-10pt}
    \resizebox{0.95\columnwidth}{!}
    {%
    \begin{tabular}{cccccccccc}
        \toprule
         \multirowcell{2}{ \\}  &  \multicolumn{9}{c}{Distribution Match ($\gamma_l = \gamma_u$)} \\
         \cmidrule(lr){2-10} 
         & \multicolumn{3}{c}{Tr-ACC} &  \multicolumn{3}{c}{Tr-bACC} & \multicolumn{3}{c}{In-bACC} \\
         \cmidrule(lr){2-4} \cmidrule(lr){5-7} \cmidrule(lr){8-10} 
         Method & All & Old & New & All & Old & New & All & Old & New \\
         \midrule
          $k$-means &  40.1 & 39.6 & 40.6 & 34.2 &  35.0 & 33.4 &   30.3&  32.9&   27.6 \\
         $\text{ORCA}^{\dagger}$~\cite{cao2021open} &  51.2& 64.9& 43.9& 25.0&  31.5& 18.6& 29.5& 39.1&  19.9  \\
         GCD~\cite{vaze2022generalized}&  \underline{55.0} & 52.1& \textbf{57.7} & 42.3& 45.9& \underline{38.6}&  38.1& 42.8&  \underline{33.4}\\
         $\text{TRSSL}^{\dagger}$~\cite{rizve2022realistic} &  41.3&  73.3&  25.4& 33.7&  46.7&  20.6&  37.9&  \underline{53.5}& 22.4 \\
         $\text{OpenCon}^{\dagger}$~\cite{sun2022opencon}& 53.5&  \textbf{79.9}& 39.9& \underline{48.5}&  62.8&  35.2& \underline{47.7}&  \textbf{62.3}& 33.2\\ 
         PromptCAL~\cite{zhang2023promptcal} & 52.3 & 72.6 & 32.1 & 46.0 & \underline{62.9} & 29.1&  38.5& 52.6& 24.4\\
         SimGCD~\cite{wen2022simple}& 51.7 & 54.3 & 49.2 &  46.5 & 59.8 & 33.2&  37.4& 44.1& 30.8\\
         BaCon~\cite{bai2023towards}& 45.8 & 40.0 & 51.5 &38.0 & 41.9 & 34.2 & 35.9&40.5& 31.2\\
         \midrule
        \rowcolor{cyan!10} Ours & \textbf{65.3} & \underline{77.4} & \underline{53.3} &\textbf{53.7} & \textbf{68.4} & \textbf{39.1} &  \textbf{48.1}&  52.9&  \textbf{43.2}\\
         \bottomrule
    
        \toprule
         \multirowcell{2}{\\} & \multicolumn{9}{c}{Distribution Mismatch ($\gamma_l \neq \gamma_u$)} \\
         \cmidrule(lr){2-10} 
         & \multicolumn{3}{c}{Tr-ACC} & \multicolumn{3}{c}{Tr-bACC} & \multicolumn{3}{c}{In-bACC} \\
         \cmidrule(lr){2-4} \cmidrule(lr){5-7} \cmidrule(lr){8-10} 
         Method & All & Old & New & All & Old & New & All & Old & New  \\
         \midrule
          $k$-means &  46.0 & 48.4 & 43.6 &41.8 & 48.4 & 35.2 & 36.9&  36.9& 37.0 \\
         $\text{ORCA}^{\dagger}$~\cite{cao2021open} & 48.8&  35.5&  55.5& 23.8&  25.5&  22.2& 27.2&  30.5& 23.8  \\
         GCD~\cite{vaze2022generalized}  &  52.8 & 56.8 & 48.9 & 44.3 & 59.7 & 28.9 & 44.6& 54.0& 35.1\\
         $\text{TRSSL}^{\dagger}$~\cite{rizve2022realistic} &    34.5&  39.0&  32.3& 31.7&  36.6&  26.8&  35.4&  39.6& 31.2 \\
         $\text{OpenCon}^{\dagger}$~\cite{sun2022opencon}&49.6&  50.7&  49.0&46.3&  51.1& \underline{41.5}& 47.4&  54.3& \underline{40.4} \\ 
         PromptCAL~\cite{zhang2023promptcal}&56.6 & \textbf{76.0} & 37.3 &54.2 & \textbf{78.0} & 30.4 & 48.1&  \textbf{67.4}& 28.8\\
         SimGCD~\cite{wen2022simple}&  \underline{65.8} & \underline{75.2} & \underline{56.4} & \underline{55.2} & \underline{77.0} & 33.4 & \underline{50.3}&  \underline{65.3}& 35.4\\
         BaCon~\cite{bai2023towards}&  56.0 & 56.5 & 55.6 &46.4 & 61.2 & 31.7 & 42.8& 50.9& 34.8\\
         \midrule
        \rowcolor{cyan!10} Ours & \textbf{66.6} & 74.2 & \textbf{59.0} &\textbf{57.3} & 68.7 & \textbf{45.9} &  \textbf{53.1}&  64.3& \textbf{41.8}\\
         \bottomrule
         
    \end{tabular}
    }
    \label{tab:main_ex2}
    \vspace{-20pt}
\end{table}

We compare our method with the state-of-the-art OWSSL methods
~\cite{cao2021open,vaze2022generalized,rizve2022realistic,sun2022opencon,zhang2023promptcal,wen2022simple,bai2023towards}.
We report the results on CIFAR-100-LT with an imbalance ratio 
 $\gamma=100$ in \tabref{tab:main_ex2}.
We explore two scenarios with different class priors of the unlabeled dataset $D_u$:
1) the class prior of $D_u$ is consistent with $D_l$, \ie, MCAR ($\gamma_l = \gamma_u$; \figref{fig:1_a} left), and
2) the class prior of $D_u$ is reversed from $D_l$, leading to a discrepancy in class prior distribution between them, \ie, MNAR ($\gamma_l \neq \gamma_u$; \figref{fig:1_a} right).
In most cases, our method outperforms others
in terms of overall accuracy for both transductive and inductive inferences.
Specifically, our method shows superior novel class accuracy, demonstrating that the density-based approach is effective in compensating for the difficulty of learning novel classes.

%% file: sec_workshop_camera/workshop_suppl_ver2.tex
\clearpage
\appendix
\numberwithin{table}{section}
\numberwithin{figure}{section}
\numberwithin{equation}{section}
\setcounter{page}{1}
\onecolumn
\section*{\Large{Appendix}}

\section{Related Works}
\label{sec:related}
\begin{table*}[!h]
\caption{Comparison of ROWSSL with other related settings.}
    \centering
    \begin{threeparttable}
    {%
    \resizebox{\textwidth}{!}{
    \begin{tabular}{cccc@{\hskip 0.1in}ccc}
        \toprule

         \textbf{Setting} & Known Classes & Novel Classes & Data Distribution & Distribution Mismatch  & Evaluation \\ \midrule
         SSL & Classify & Not present & Balanced & \xmark & Inductive \\ 
         Robust SSL \cite{guo2020safe} & Classify & Reject & Balanced& \xmark &  Inductive \\ 
         LT-SSL \cite{wei2021crest} & Classify & Not present& Imbalanced& \xmark &  Inductive\\
         RLT-SSL \cite{wei2023towards} & Classify & Not present& Imbalanced& \checkmark &  Inductive\\
         NCD \cite{hsu2019multi} & Not present & Discover& Balanced& - &  Transductive \\ 
                DA-NCD \cite{yang2023bootstrap} & Not present & Discover& Imbalanced& - &  Transductive \\ 
         OWSSL/GCD \cite{vaze2022generalized,cao2021open} & Classify & Discover& Balanced& \xmark &  Transductive \\ 
         DA-GCD \cite{bai2023towards} & Classify & Discover& Imbalanced& \xmark &  Transductive \tnote{*} \\ \midrule
        \rowcolor{cyan!10} \textbf{ROWSSL} & Classify & Discover \& Classify \tnote{**} & Imbalanced&  \checkmark & 
         Transductive \& Inductive\\
    
         \bottomrule
         
    \end{tabular}

    }}
        \footnotesize
\begin{tablenotes}
\begin{scriptsize}
\item[*] Evaluated on the disjoint test dataset, but it requires to see the entire test dataset for inference, i.e., transductive inference.
\item[**] Discover novel classes on the unlabeled training dataset and classify them on the disjoint test dataset.%
\end{scriptsize}
\end{tablenotes}
    \end{threeparttable}
    \label{tab:setting}
\end{table*}

\paragraph {Open-world semi-supervised learning (OWSSL)}
or generalized category discovery (GCD) is a transductive learning setting which extends semi-supervised learning (SSL) and novel category discovery (NCD)~\cite{hsu2015neural} by classifying known classes as well as discovering novel classes in the unlabeled training dataset.
Vaze et al.~\cite{vaze2022generalized} addresses this task via contrastive learning (CL) on a pre-trained vision transformer (ViT) \cite{dosovitskiy2020image,caron2021emerging} followed by constrained $k$-means clustering \cite{macqueen1967some}.
Since then,
a plethora of works have explored CL to achieve robust representations in OWSSL.
XCon \cite{fei2022xcon} learns fine-grained discriminative features by dataset partitioning.
PromptCAL \cite{zhang2023promptcal}, DCCL \cite{pu2023dynamic}, OpenNCD \cite{liu2023open}, and CiPR \cite{hao2023cipr} construct an affinity graph, and OpenCon \cite{sun2022opencon} utilizes a prototype-based novelty detection strategy to mine reliable positive pairs for the contrastive loss.
GPC \cite{zhao2023learning} introduces a novel representation learning strategy based on a semi-supervised variant of the Gaussian mixture model.
SPTNet~\cite{wang2024sptnet} proposes an iterative optimization method which optimizes both model and data parameters.
In parallel with them, ORCA~\cite{cao2021open}, NACH~\cite{guo2022robust}, and OpenLDN~\cite{rizve2022openldn} utilize pairwise learning, generating pseudo-labels for unlabeled data by ranking distances in the feature space.
ORCA and NACH also propose uncertainty-based loss to alleviate known class bias caused by different learning speeds between known and novel classes.

However, these advances are mostly based on the assumption that the class prior of the training dataset is balanced; indeed, data imbalance poses further challenges in OWSSL.
For example, while a majority of methods proposed for OWSSL employed CL,
it has been known that CL is not immune to data imbalance, such that representations learned on long-tailed distribution might be biased toward head classes~\cite{jiang2021self}.
Also, they mostly rely on $k$-means clustering,
which assumes the presence of isotropic data clusters~\cite{liang2012k,wu2009adapting}, such that the uniform cluster prior assumption often hampers representation learning~\cite{assran2022hidden}.
In the case of pairwise learning-based methods,
the classifier is learned to be biased toward head classes due to the lack of positive pairs in tail classes~\cite{chuyu2023novel}.
Learning pace-based methods
only take account of the uncertainty of known and novel classes, such that it might be difficult to distinguish between known-tail classes and novel-head classes~\cite{wang2022partial}.
Lastly, non-parametric methods~\cite{vaze2022generalized,zhang2023promptcal,bai2023towards} apply $k$-means clustering at inference time, which requires access to the entire test dataset for inference, often unattainable in real-world scenarios and hinders online inference, i.e., inductive learning. 

In this paper, we advance OWSSL to a more practical setting, considering long-tailed distribution with class prior distribution mismatch, and inductive inference.
Also, we address the aforementioned problems by density estimation on the latent feature space to achieve balanced CL and reduce the classifier bias toward the head and known classes.

\paragraph {Long-tailed recognition} 
considers imbalanced class prior, which is natural in real-world scenarios.
Early approaches to combat the imbalance include data re-sampling \cite{chawla2002smote}, re-weighting \cite{cui2019classbalanced}, and margin-based approach \cite{cao2019learning} with respect to given class-wise sample sizes.
Based on this, DARP~\cite{kim2020distribution} and CReST~\cite{wei2021crest} introduce long-tailed semi-supervised learning methods utilizing distribution alignment.
Recently, ACR~\cite{wei2023towards} and PRG~\cite{duan2023towards} suggest realistic long-tailed semi-supervised learning setting considering class prior distribution mismatch between labeled and unlabeled datasets, i.e., MNAR. 
However, their closed-world assumption hinders direct application to ROWSSL.
On the other hand, self-supervised learning under long-tailed distribution has also been investigated \cite{jiang2021self,kukleva2023temperature}.
As the temperature parameter plays a significant role in shaping the learning dynamics of CL~\cite{wang2020understanding,wang2021understanding}, Kukleva et al.~\cite{kukleva2023temperature} adjusts the temperature parameter with cosine scheduling to improve linear separability between head and tail classes.
Different from prior works, our DTS dynamically adjusts the temperature of the contrastive loss based on the estimated density rather than predefined cosine scheduling.

BaCon~\cite{bai2023towards} proposed distribution-agnostic GCD,
but their setting is different from ours in that
1) it performs transductive inference on the test set by $k$-means clustering on the entire test dataset for evaluation, and
2) it does not assume the potential class prior distribution mismatch.
Our density-based approach effectively addresses ROWSSL, outperforming BaCon in both inductive and transductive learning settings.

\section{Implementation Details}

Our algorithm is developed using PyTorch~\cite{paszke2019pytorch} and we conduct all the experiments with a single NVIDIA RTX A5000 GPU. 
Following~\cite{vaze2022generalized,wen2022simple}, 
 all methods are trained with a ViT-B/16~\cite{dosovitskiy2020image} backbone with DINO~\cite{caron2021emerging} pre-trained weights, and use the output [CLS] token with a dimension of 768 as the feature representation, and the MLP $g$ projects the feature representation to a 256-dimensional vector. 
All methods were trained for 200 epochs with a batch size of 128, and we fine-tuned the final transformer block using standard SGD with momentum 0.9, and an initial learning rate of 0.1 which we decay with a cosine annealed schedule. %
The balancing factor $\lambda_{\text{rep}}$ for the contrastive loss is set to 0.35.
For the classification objective, we set $\tau_s$ to 0.1, and $\tau_t$ is initialized to 0.07, then warmed up to 0.04 with a cosine schedule in the starting 30 epochs.
The weight of the mean-max regularization $\varepsilon$ is set to 4.
We set the number of tailedness prototype $M$ to be the same as the total class number $C$, with the moving average factor $\lambda_{\text{tail}}$ to 0.9.
The queue size $Q$ is set to 4096
and the $K$-nearest neighbor distance is computed on $K=15$.
The default hyperparameters 
$\tau_{\text{min}}$ and $\tau_{\text{max}}$ for dynamic temperature scaling are set to 0.05 and 1, and we set $\lambda_{\text{var}}$ to 1. %

\vspace{-2pt}

\section{Illustration of Proposed Framework}

\begin{figure*}[h]
\centering
{\includegraphics[width=0.99\linewidth]{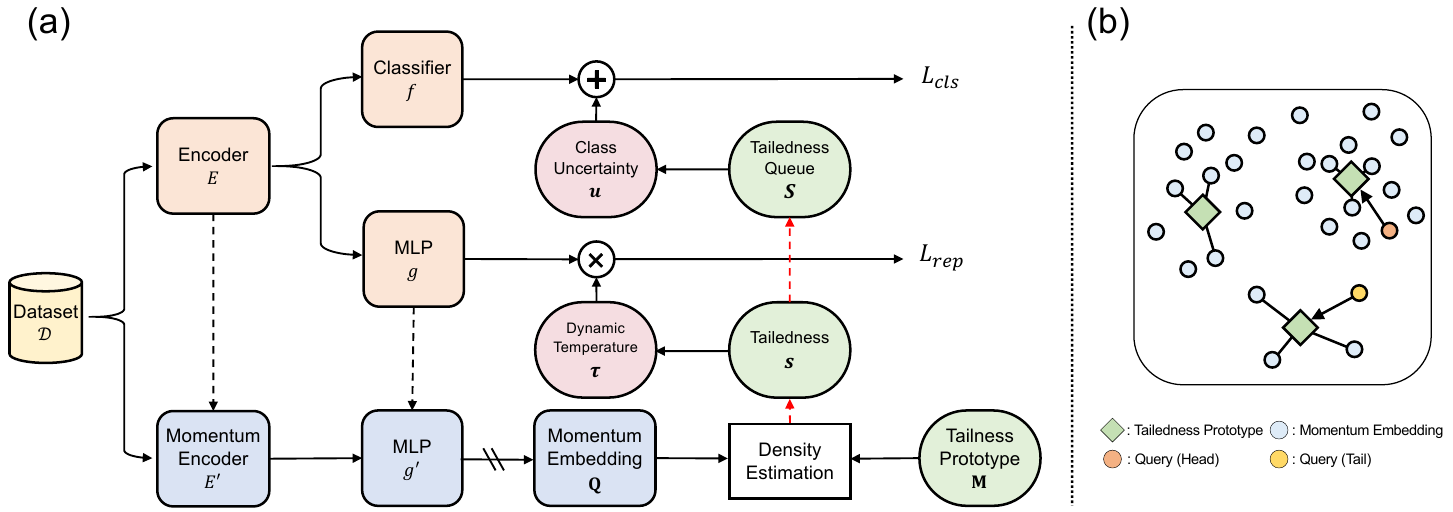}}\\
\caption{
(\textbf{a}) Overall framework of the proposed DTS.
``\textbackslash \textbackslash'' stands for \textit{stop gradient}.
(\textbf{b}) Example of tailedness estimation.
}
\label{fig:2}
\end{figure*}

\section{More Experiments and Discussions}

We present ablation studies to evaluate and understand the contribution of each component of our method. 
To examine how the performance is influenced by the long-tail distribution, we categorize known and novel classes into three separate groups: \{Many, Median, Few\}, based on the number of data per class.
All experiments are conducted in MCAR or the distribution matched setting.

\subsection{Design choices for the tailedness estimation.}

\begin{figure}[!htb]
   \begin{minipage}{0.48\textwidth}
     \centering
     \includegraphics[width=0.9\columnwidth]{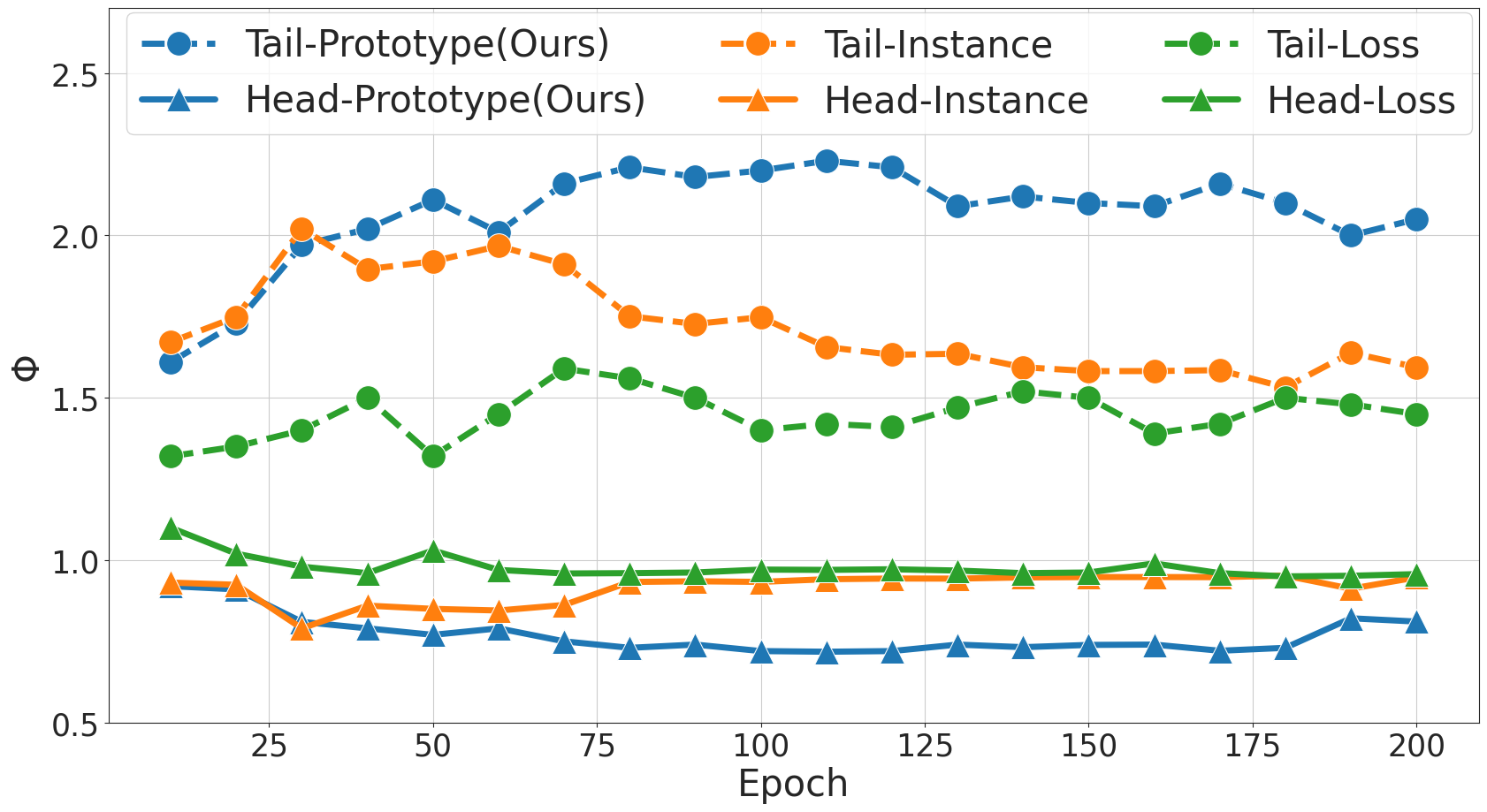}
     \caption{Comparison of tail discovery methods.}
     \label{fig:tail_a}
   \end{minipage}\hfill
   \begin{minipage}{0.48\textwidth}
     \centering
     \includegraphics[width=0.9\columnwidth]{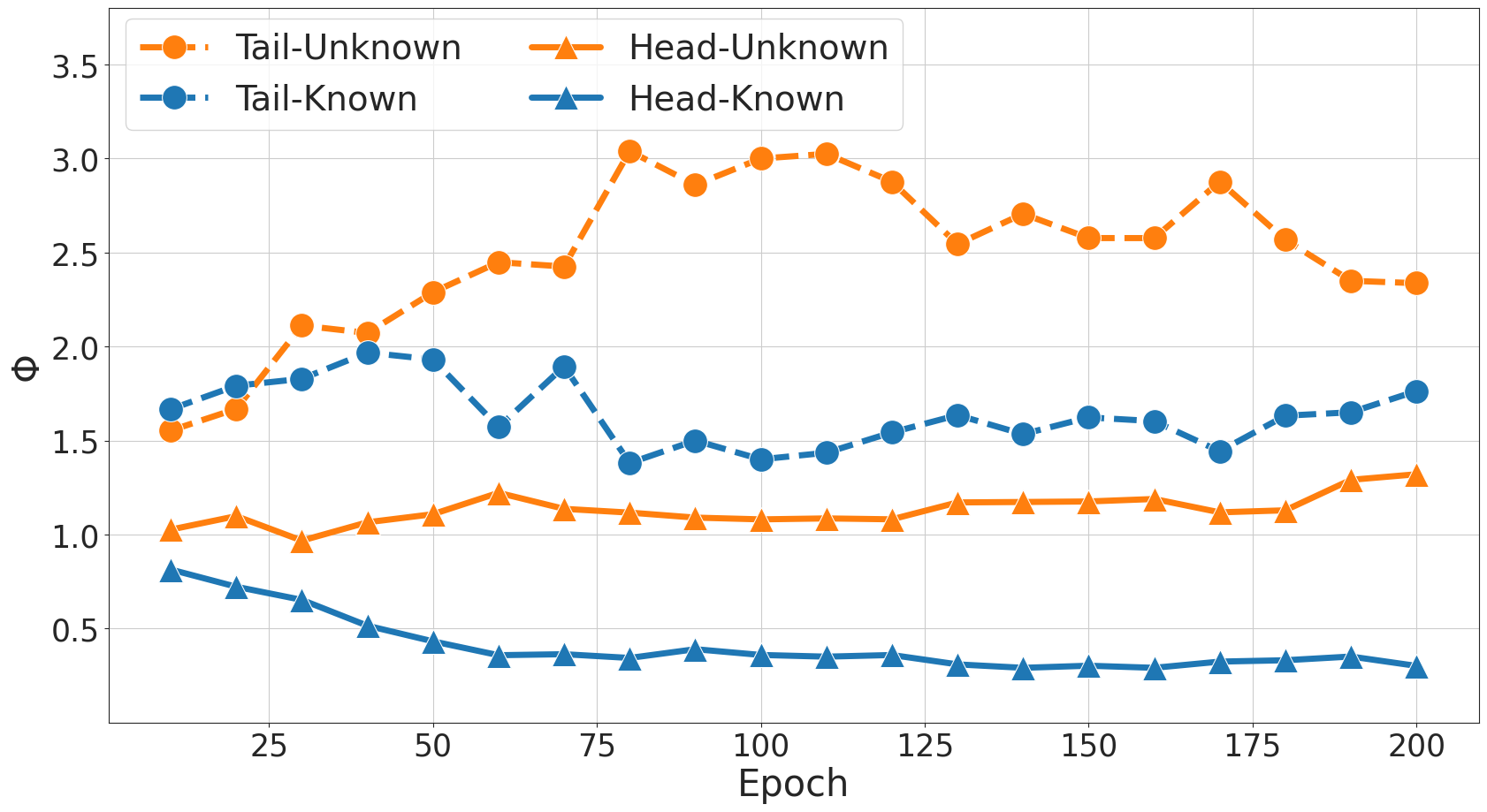}
     \caption{Comparison between known and novel classes of our method.}
     \label{fig:tail_b}
   \end{minipage}
\end{figure}

In this experiment,
we measure the correlation metric used in~\cite{bai2023effectiveness,zhou2022contrastive} to validate our tailedness estimation method.
 Specifically, we divide the training dataset into two groups $\mathcal{T} \in \{\text{Head}, \text{Tail}\}$, and compute 
 $\phi=\frac{|\mathcal{T} \cap \mathcal{X}^{\text{sub}}|/|\mathcal{X}^{\text{sub}}|}{|\mathcal{T} \cap \mathcal{X}|/|\mathcal{X}|}$ for each group,
 where
 $\mathcal{X}$ is the whole dataset and
 $\mathcal{X}^{\text{sub}}$ is the subset of samples which have the top-10\% lowest density.
The parameter $\phi$ serves as an indicator of the ability to identify tail samples:
when the target group is head (tail), lower (higher) $\phi$ indicates that the method effectively localizes tail samples.
In~\figref{fig:tail_a},
we compare our prototype-based estimation (``Prototype'') with the instance-wise estimation baseline (``Instance'') and the loss-based estimation (``Loss'')~\cite{zhou2022contrastive}.
As shown in \figref{fig:tail_a}, our method discovers tail samples more effectively than others.
To observe the effect of our method on discriminating known and novel classes, we further divided head and tail groups into known and novel classes.
In~\figref{fig:tail_b}, 
we observe that the difference between known and novel classes is not captured well at the beginning of training, but samples from novel classes begin to exhibit larger tailedness in the representation space than those from known classes through training.
This implies that the learning difficulty could be captured by our density-based approach, which is further discussed in the following.

\subsection{Design choices for the temperature in CL}
To validate the effectiveness of the dynamic temperature for the contrastive loss%
, we experiment with different choices of the temperature. %
In addition to the constant temperature ($\tau=0.07$), we compare our density-based approach with the estimated class prior by hard pseudo-labels for the classifier following~\cite{yang2023bootstrap} as baselines,
and the true class prior as an oracle,
where we apply the same min-max normalization as our method. %
We observe that our proposed method achieves better overall accuracy compared to baselines.
Interestingly, our method even outperforms the oracle with the true class prior, implying that the learning difficulty of classes is not strictly proportional to the  %
class prior, and our density-based approach can be more effective in addressing it.

\subsection{Design choices for the class uncertainty}
In this experiment, we validate the effectiveness of the choice of $\mathbf{u}$, which is the standard deviation of class-wise tailedness scores.
We compare the variance of the maximum softmax probability as confidence for each class and the estimated distribution~\cite{yang2023bootstrap} as baselines and the ground-truth class prior as an oracle.
For both estimated and ground-truth class prior, we convert the class frequency into a normalized probability distribution.
As shown in \tabref{tab:ucr}, our method achieves comparable performance to the oracle performance.
Notably, our method boosts performance by 6.2\% in novel classes and 3.8\% in tail classes.
This result confirms that focusing on class uncertainty is more effective than using class prior for mitigating the bias of the classifier in the ROWSSL setting.

\begin{table}[h]
\begin{minipage}{.5\textwidth}
    \centering
    \caption{Ablation study on $\tau$.}
    \scalebox{0.72}{
    \begin{tabular}{ccccccc}
        \toprule
         &  \multicolumn{6}{c}{CIFAR-100-LT} \\
         \cmidrule(lr){2-7}
         Method &  All & Old & New & Many & Med. & Few \\
         \midrule
         Constant &  45.3& 55.1& 35.5&  \underline{62.1}&  53.8& 20.0 \\
         Estimated prior & 46.8& \textbf{59.2} & 34.4&  61.9& \underline{55.2}& 23.3 \\
         True prior &  \underline{47.2}& \underline{57.4}& \underline{36.8}&  \textbf{62.2}&  54.8& \underline{24.3} \\
         \midrule
       \rowcolor{cyan!10}  Ours &  \textbf{48.1} & 52.9 & \textbf{43.2} & 59.2 & \textbf{58.0} & \textbf{27.7} \\
         \bottomrule
    \end{tabular}}
    \label{tab:abl_temp}
\end{minipage}
\begin{minipage}{.5\textwidth}
    \centering
    \caption{Ablation study on $\mathbf{u}$.}
    \scalebox{0.72}{
    \begin{tabular}{ccccccc}
        \toprule
         &  \multicolumn{6}{c}{CIFAR-100-LT} \\
         \cmidrule(lr){2-7}
         Method &  All & Old & New & Many & Med. & Few \\
         \midrule
         Confidence &  43.3& 48.4& \underline{38.1}&  58.8& 50.2& 20.7 \\
         Estimated prior & 45.5& \underline{54.9}& 36.1&  \underline{60.1} & 54.8& 21.6 \\
         True prior &  \underline{47.9} & \textbf{58.7} & 37.0& \textbf{62.9} & \underline{56.7} & \underline{23.9} \\
         \midrule
        \rowcolor{cyan!10} Ours & \textbf{48.1} & 52.9 & \textbf{43.2}& 59.2 & \textbf{58.0} & \textbf{27.7} \\
         \bottomrule
    \end{tabular}}
    \label{tab:ucr}
\end{minipage}
\end{table}

\clearpage

\subsection{Contribution of each component}
We examine the impact of each component in~\tabref{tab:component}.
 Specifically, starting from the baseline~\cite{wen2022simple}, we ablate
 the momentum encoder~\cite{he2020momentum} and 
 dynamic temperature scaling and class uncertainty-aware pseudo-labeling.%
Comparing experiments (b) and (c), the proposed dynamic temperature scaling improves performance by 2.7\% and 1.4\% for head classes, alongside 6.1\% and 14\% for tail classes on the CIFAR-100-LT and CUB-200-LT datasets, respectively.
This indicates that our method learns discriminative semantic structures for both head and tail classes.
From (b) and (d), the proposed class uncertainty-aware pseudo-labeling yields a notable improvement in all metrics.
Specifically, introducing $\mathbf{u}$ enhances performance by 7.2\% and 8.4\% in novel classes, with 5.6\% and 11.0\% in tail classes for each dataset, effectively mitigating classification bias towards known and head classes.
The full version of our method (e) shows superior performance on all evaluation metrics, which experimentally demonstrates that our approach plays a crucial role in addressing ROWSSL.

\begin{table}[h]
    \centering
    \caption{Component analysis of DTS.}
     \resizebox{0.7 \columnwidth}{!}
    {%
    \begin{tabular}{c|ccc|cccccc}
        \toprule
         \multirow{2}{*}{Index}& \multicolumn{3}{c|}{Component} & \multicolumn{6}{c}{CIFAR-100-LT} \\
         \cmidrule(lr){2-4} \cmidrule(lr){5-10}
         & Momentum & Dynamic $\tau$ & Uncertainty \textbf{u} & All & Old & New & Many & Med. & Few \\
         \midrule
         (a) & \xmark & \xmark & \xmark & 38.8 & 50.9 & 26.7 & 55.9 & 49.9 & 9.3 \\
         (b) & \checkmark & \xmark & \xmark & 41.5 & \underline{56.5} & 28.5 & 56.4 & 52.7 & 14.4 \\
         (c) & \checkmark & \checkmark & \xmark & 45.7 & 55.5 & \underline{35.9} & 59.1 & \underline{56.1} & \underline{20.5} \\
         (d) & \checkmark & \xmark & \checkmark & \underline{47.6} & \textbf{59.6} & 35.7 & \textbf{68.5} & 55.6 & 20.0 \\
         \midrule
        \rowcolor{cyan!10} (e) & \checkmark & \checkmark & \checkmark & \textbf{48.1} & 52.9 & \textbf{43.2} & \underline{59.2} & \textbf{58.0} & \textbf{27.7} \\
         \bottomrule
    \end{tabular}}
    \label{tab:component}
\end{table}

\subsection {Unknown class numbers}
In real-world applications, we often do not have prior knowledge of the true number of classes $C$.
In~\tabref{tab:unknown}, we estimate the number of classes $\widehat{C}$ and use it for evaluation depending on the type of methods:
for non-parametric clustering-based methods \cite{vaze2022generalized,bai2023towards}, we apply Brent's algorithm to estimate $\widehat{C}$ as in \cite{vaze2022generalized}, and
for parametric classification methods~\cite{cao2021open,wen2022simple} and ours, we provide an arbitrarily large number, \eg, $\widehat{C}_\text{init} = 2C$, and estimate $\widehat{C}$ by eliminating inactivated classes, \ie, classes without mappings from any training data.
Notably, the uniform prior assumption in the $k$-means algorithm leads GCD and BaCon to significantly underestimate the total class number in long-tailed datasets, resulting in overall performance degradation.
In the case of ORCA,
its pairwise learning could be dominated by known and head classes as pairs mostly consist of data from known and head classes, and
its binary uncertainty estimation would not be suitable for distinguishing known-tail and novel-head classes,
resulting in significant inactivation of classification heads.
Our method demonstrates comparable performance to scenarios where the number of classes is known, with only a 1.0\% decrease in overall inductive accuracy.

\begin{table}[h] 
    \centering
    \caption{Comparison results on CIFAR-100-LT ($\gamma_l = \gamma_u$) with an unknown number of classes.}
    \scalebox{0.72}{
    {%
    \begin{tabular}{cccccccccccc}
        \toprule
         & &  & \multicolumn{3}{c}{Tr-ACC} & \multicolumn{3}{c}{Tr-bACC} & \multicolumn{3}{c}{In-bACC} \\
         \cmidrule(lr){4-6}  \cmidrule(lr){7-9} \cmidrule(lr){10-12} 
         Method & Param. & Est. $\widehat{C}$ &  All & Old & New & All & Old & New & All & Old & New \\
         \midrule
         ORCA  & \cmark & 59& 46.9 & 50.7 & \underline{45.1} &25.2& 31.5& 19.0& 28.1& 37.7& 18.6\\
         GCD   & \xmark &76 & 44.7 &47.4 & 39.3& 37.9& 38.1& \underline{37.7} &38.6&51.6 &25.7 \\
         SimGCD &\cmark &145& \underline{52.8} & \textbf{73.2} & 42.6& \underline{42.9}& \underline{55.6} & 30.3& \underline{41.6}& \underline{56.0}& 27.2\\
         BaCon   & \xmark &79 & 48.4 & 63.1 & 36.0&42.4 & 52.3 & 32.5 & 33.1& 33.5& \underline{32.7} \\
         \midrule
      \rowcolor{cyan!10}   Ours & \cmark&94& \textbf{60.8} & \underline{72.0} & \textbf{49.6} &\textbf{51.3} & \textbf{61.2} & \textbf{41.4}& \textbf{47.1} & \textbf{57.6} & \textbf{36.6} \\
         
         \bottomrule         
    \end{tabular}}}   
    \label{tab:unknown}
\end{table}

\subsection{Number of tailedness prototypes}
To evaluate the performance sensitivity in relation to the number of tailedness prototypes $M$,
we conduct an ablation study on different prototype numbers.
As shown in~\tabref{tab:pro},
aligning the number of prototypes with the class number yields the best performance.
In general, our method demonstrates robustness across various numbers of prototypes, yielding the best performance among compared methods in most cases.
Note that matching the number of prototypes with the true number of classes might not always result in the best performance,
because multiple fine-grained classes might form a single coarse-grained class or a class might consist of multiple local clusters~\cite{qian2019softtriple}.
\begin{table}[h]
    \centering
     \caption{Comparison results on CIFAR-100-LT ($\gamma_l = \gamma_u$) with various number of prototypes.}
     \scalebox{0.82}{
    {%
    \begin{tabular}{c|ccccccccc}
        \toprule
       \multicolumn{1}{c}{}
       & \multicolumn{3}{c}{Tr-ACC} &   \multicolumn{3}{c}{Tr-bACC} & \multicolumn{3}{c}{In-bACC} \\
         \cmidrule(lr){2-4}  \cmidrule(lr){5-7} 
         \cmidrule(lr){8-10}
         $M$ & All & Old & New & All & Old & New & All & Old & New\\
         \midrule
         50  & 60.2 & 74.3 & 46.1& 49.0 & 59.8  &  38.2 & 46.2 & 53.0  & 39.4 \\
         200  & \underline{63.4} & 74.7 & 52.1 & \underline{51.2} & \underline{64.6} & 37.8 & \underline{47.7} & \underline{54.0} & \underline{41.5}   \\
         300   & 63.0& \underline{75.0}& \underline{51.0}&48.5 & 57.5  & \textbf{39.6} & 47.4 & \textbf{54.8} & 40.1  \\
         \midrule
       \rowcolor{cyan!10}  100 & \textbf{65.3} & \textbf{77.4} & \textbf{53.3} &\textbf{53.7} & \textbf{68.4} & \underline{39.1}  & \textbf{48.1}& 52.9 & \textbf{43.2}    \\
         \bottomrule
         
    \end{tabular}}}
  
     \label{tab:pro} 
\end{table}

\subsection{Results with different imbalance ratios}
In previous experiments, we use $\gamma = 100$ for CIFAR-100-LT.
In~\Crefrange{tab:cifar_bal}{tab:cifar_diff_imb},
we conduct an ablation study for different imbalance ratios ($\gamma = 1,10$) on CIFAR-100-LT.
Our method shows superior performance in overall
accuracy
for various imbalance ratios, showing its generalization ability for the different class priors.
Bacon~\cite{bai2023towards} often outperforms our method in novel class accuracy in transductive inference, however, its performance is degraded in inductive inference, while our method maintains good performance in inductive inference.

\begin{table*}[h]
    \centering
    \caption{Results on balanced CIFAR-100 ($\gamma = 1$).}
    \resizebox{0.55 \textwidth}{!}
    { 
    \begin{tabular}{cccccccccc}
        \toprule
         & \multicolumn{3}{c}{Tr-ACC} & \multicolumn{3}{c}{Tr-bACC} & \multicolumn{3}{c}{In-bACC} \\
         \cmidrule(lr){2-4} \cmidrule(lr){5-7} \cmidrule(lr){8-10} 
         Method & All & Old & New & All & Old & New & All & Old & New\\
         \midrule
         $k$-means & 49.2 & 50.1 & 48.5 & 49.3 & 50.1 & 48.6 & 50.0 & 54.9 & 45.2 \\
        $\text{ORCA}^{\dagger}$ & 41.6 & 50.1 & 37.3 & 43.7 & 50.1 & 37.3 & 44.5 & 52.5 & 36.5 \\ 
        GCD & 64.6 & \underline{72.7} & 60.5 & 66.6 & \underline{72.8} & 60.3 & 63.5 & 74.9 & 52.1 \\ 
        $\text{TRSSL}^{\dagger}$ & 50.3 & 71.0 & 40.0 & 55.5 & 71.0 & 40.1 & 66.3 & \textbf{83.1} & 49.5 \\ 
        SimGCD & \underline{65.4} & 71.9 & \underline{62.6} & \underline{67.2} & 71.9 & \underline{62.6} & \underline{69.8} & 77.3 & \underline{62.4} \\
        BaCon & 65.3 & 72.3 & 61.8 & 67.0 & 72.4 & 61.7 & 69.2 & 81.0 & 57.4 \\
        \midrule
      \rowcolor{cyan!10}  Ours & \textbf{69.0} & \textbf{79.4} & \textbf{63.8} & \textbf{71.6} & \textbf{79.4} & \textbf{63.8} & \textbf{72.6} & \underline{81.9} & \textbf{63.2} \\
         \bottomrule
         
    \end{tabular}

    }
    \label{tab:cifar_bal}
\end{table*}

\begin{table*}[h]
    \centering
    \caption{Results on CIFAR-100-LT with $\gamma = 10$.}
    \resizebox{\textwidth}{!}
    { 
    \begin{tabular}{cccccccccc|ccccccccc}
        \toprule
         & \multicolumn{9}{c|}{Distribution Match ($\gamma_l = \gamma_u$)}& \multicolumn{9}{c}{Distribution Mismatch ($\gamma_l \neq \gamma_u$)} \\ 
         
         \cmidrule(lr){2-10} \cmidrule(lr){11-19}
         & \multicolumn{3}{c}{Tr-ACC} & \multicolumn{3}{c}{Tr-bACC} & \multicolumn{3}{c|}{In-bACC} & \multicolumn{3}{c}{Tr-ACC} & \multicolumn{3}{c}{Tr-bACC} & \multicolumn{3}{c}{In-bACC}\\
         \cmidrule(lr){2-4} \cmidrule(lr){5-7} \cmidrule(lr){8-10} \cmidrule(lr){11-13} \cmidrule(lr){14-16} \cmidrule(lr){17-19} 
         Method & All & Old & New & All & Old & New & All & Old & New & All & Old & New & All & Old & New & All & Old & New \\
         \midrule
         $k$-means& 46.7 & 44.0 & 47.9 & 41.6 & 38.6 & 44.6 & 41.7 & 43.5 & 40.0 & 51.2 & 55.2 & 49.2& 48.7 & 53.6 & 43.8 & 48.7 & 55.5 & 42.0 \\ 
        $\text{ORCA}^{\dagger}$ & 44.2 & 50.8 & 40.9 & 34.3 & 41.4 & 27.3 & 39.2 & 51.2 & 27.3 &40.6 & 43.0 & 39.3 &30.7 & 36.8 & 24.6 & 31.9 & 39.3 & 24.4 \\ 
        GCD &55.5 & 61.2 & 52.8 & 52.5 & 60.4 & 44.7 & 51.9 & 63.5 & 40.3 & 60.6 & \underline{75.1} & 53.4&58.8 & \underline{71.8} & 45.8 & 56.3 & 71.5 & 41.1  \\ 
        $\text{TRSSL}^{\dagger}$ & 42.9 & \underline{66.1} & 31.4 & 42.4 & 56.1 & 28.7 & 49.0 & \underline{65.7} & 32.3 & 43.3 & 59.6 & 35.2&43.7 & 54.7 & 32.7 & 51.4 & 63.4 & 39.4 \\ 
        SimGCD & 54.2 & 59.1 & 51.8 & 53.4 & \underline{62.2} & 44.6 & 52.8 & 61.5 & 44.2 & 62.1 & 73.7 & \underline{56.2} &\underline{59.4} & 69.6 & 49.2 & \underline{62.8} & \underline{75.8} & \underline{49.9}\\ 
        BaCon & \underline{59.7} & 58.2 & \textbf{60.5} & \underline{54.5} & 55.9 & \textbf{53.2} & \underline{55.1} & 64.3 & \underline{45.9}& \textbf{63.9} & 68.4 & \textbf{61.6}&  58.7 & 68.1 & \underline{49.3} & 59.0 & 72.9 & 45.2 \\
        \midrule
     \rowcolor{cyan!10}   Ours &  \textbf{61.7} & \textbf{71.4} & \underline{55.4} & \textbf{60.8} & \textbf{70.6} & \underline{51.0} & \textbf{62.2} & \textbf{73.3} & \textbf{51.1} & \underline{63.8} & \textbf{79.7} & 55.9 &\textbf{62.9} & \textbf{73.4} & \textbf{52.4} & \textbf{64.3} & \textbf{78.0} & \textbf{50.6} \\ 
         \bottomrule
         
    \end{tabular}
    }
    \label{tab:cifar_diff_imb}
\end{table*}
\newpage

\section{Hyperparameter Analysis}
We conduct ablation experiments on critical hyperparameters of DTS, including
(1) the number of neighbors for the $K$-NN,
(2) $\tau_{max}$ for dynamic temperature scaling, and
(3) $\lambda_\text{var}$ for class-uncertainty aware pseudo labeling.
We report overall inductive balanced accuracy performance on the CIFAR-100-LT dataset with distribution matched setting ($\gamma_l = \gamma_u$).

\subsection{Number of neighbors for the $K$-NN}
We consider $K = \{5, 10, 15, 20, 25\}$ for inspecting the impact of the number of nearest neighbors on tailedness estimation.
As shown in \figref{fig:knn}, the optimal number of nearest neighbors is 15. 
When the neighborhood size is increased to include large neighbors ($K > 20$), we observe a slight degradation in performance, implying that the larger neighborhoods might accurately capture the local density that represents the class prior distribution.

\subsection{Hyperparameter $\tau_{\text{max}}$}
In \figref{fig:tau}, we investigate the effect of the range of $\tau$ by considering $\tau_{\text{max}} = \{0.5, 0.7, 0.9, 1.0,1.5\}$ with $\tau_{\text{min}}=0.05$, 
where $\tau_{\text{max}} = 1.0$ shows the best performance.
We argue that it optimally balances the uniformity and alignment of representation. 
A narrow range of tau ($\tau_{\text{max}} < 0.7$) may disrupt the semantic representation, while a wide range of tau ($\tau_{\text{max}}>1.0$) could negatively impact learning instance-specific features.

\subsection{Hyperparameter $\tau_{\text{min}}$}
In \figref{fig:tau2}, we examine the effect of the range of $\tau$ by considering $\tau_{\text{min}} = \{0.01, 0.02, 0.05, 0.1, 0.3\}$ with $\tau_{\text{max}}=1.0$, 
where $\tau_{\text{min}} = 0.05$ shows the best performance.
We argue that it optimally balances the uniformity and alignment of representation. 
A high minimum value of tau ($\tau_{\text{min}} > 0.1$) may hinder the learning instance-specific features, 
while a low minimum value of tau ($\tau_{\text{min}} < 0.05$) may disrupt the semantic representation.

\subsection{Hyperparameter $\lambda_\text{var}$}
\begin{figure}[h]
    \centering
    \begin{subfigure}[b]{0.22\textwidth}
        \includegraphics[width=\textwidth]{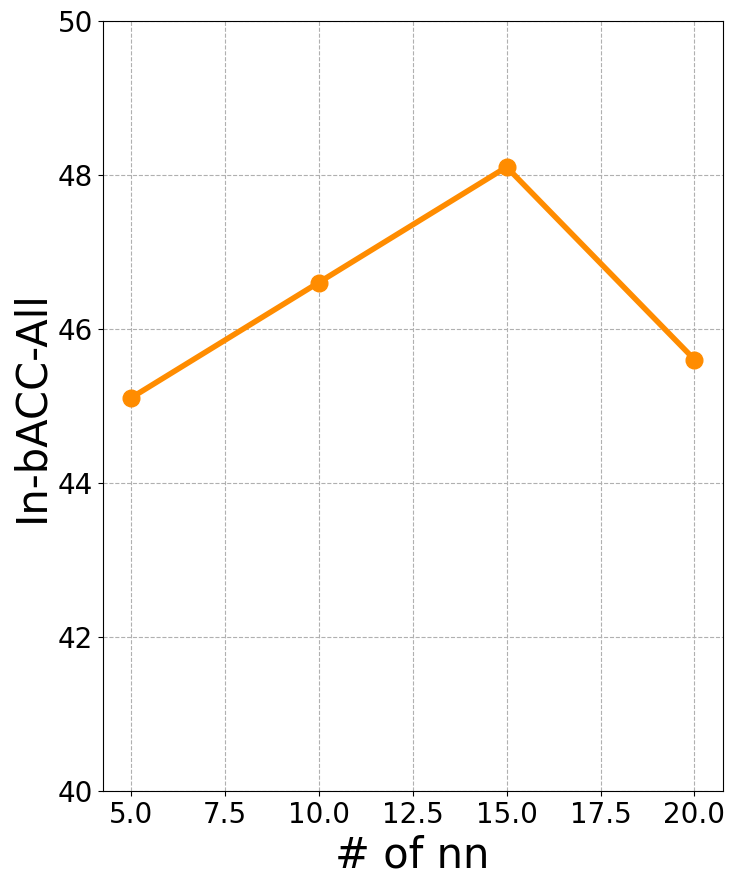}
        \caption{$K$-NN}
        \label{fig:knn}
    \end{subfigure}
    \hfill
    \begin{subfigure}[b]{0.22\textwidth}
        \includegraphics[width=\textwidth]{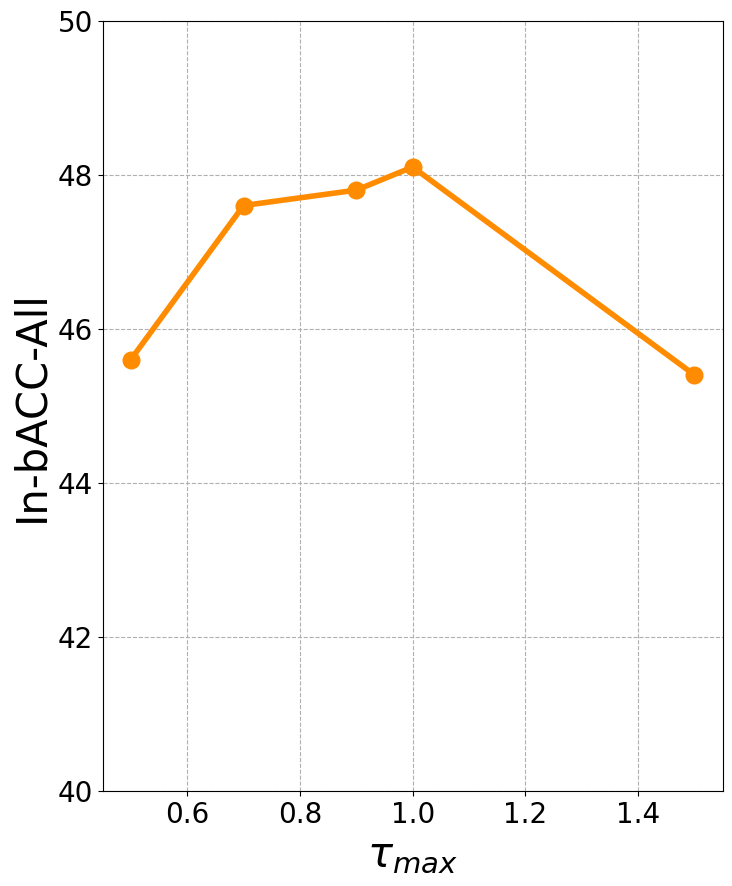}
        \caption{$\tau_{max}$}
        \label{fig:tau}
    \end{subfigure}
    \hfill
    \begin{subfigure}[b]{0.22\textwidth}
        \includegraphics[width=\textwidth]{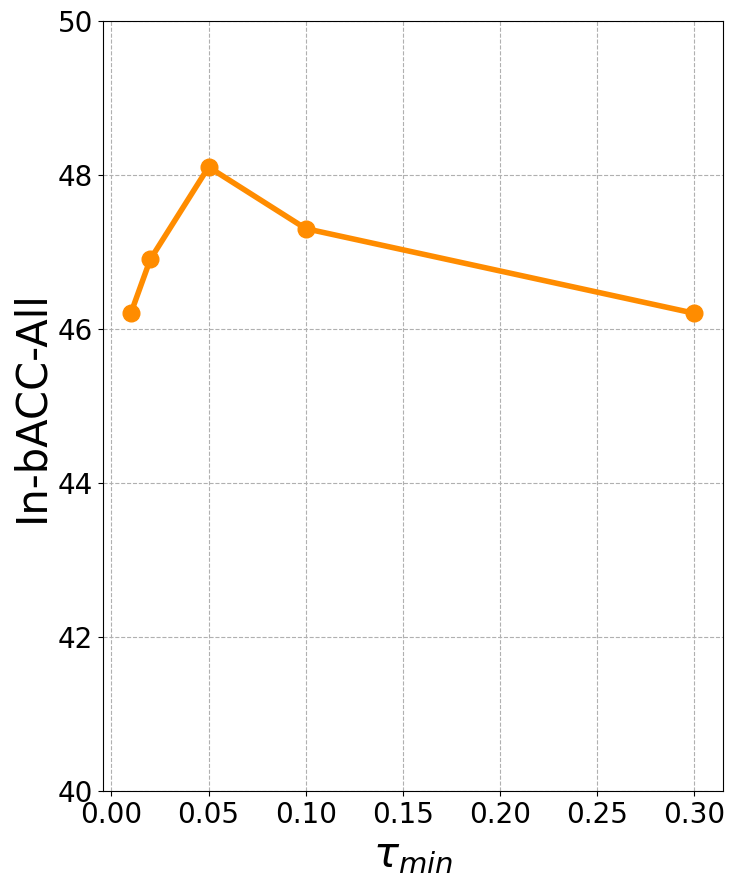}
        \caption{$\tau_{min}$}
        \label{fig:tau2}
    \end{subfigure}
    \hfill
    \begin{subfigure}[b]{0.22\textwidth}
        \includegraphics[width=\textwidth]{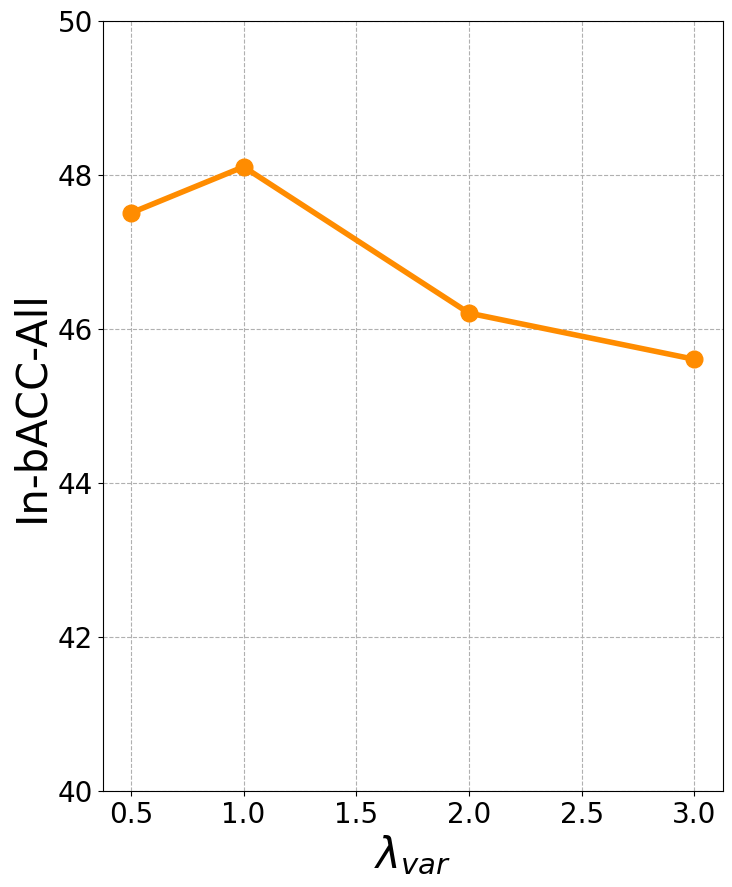}
        \caption{$\lambda_\text{var}$}
        \label{fig:var}
    \end{subfigure}
    \caption{Analysis of hyperparameters.}
\end{figure}

We examine the effect of the weight parameter $\lambda_\text{var}$ as illustrated in~\figref{fig:var}, where we consider $\lambda_\text{var}=\{0.5,1.0,2.0,3.0\}$.
Among them, $\lambda_\text{var}=1$ shows the best performance.
 Notably, a larger weight parameter appears to adversely affect the information contained in the original output logits of the cosine classifier.

\section{Detailed Results of Main Experiments}

To better examine the impact of dataset imbalance, we conduct a detailed comparison 
in~\Crefrange{tab:cifar_mcar}{tab:herb_mnar}.
In %
\tabref{tab:herb_mnar}, we report the performance in Missing Not At Random (MNAR) scenarios for the in-nature long-tailed dataset %
Herbarium19.
Our approach demonstrates a significant performance improvement for novel and tail classes,
where the conventional open-world and long-tailed learning strategies do not take into account the importance of learning tail and novel classes, respectively.
This validates that our method effectively addresses known and head class bias issues.

\begin{table*}[h]
    \centering
    \caption{Results on CIFAR-100-LT ($\gamma_l = \gamma_u$).}
    \resizebox{\textwidth}{!}
    { 
    \begin{tabular}{cccccccccccccccc}
        \toprule
         &  \multicolumn{15}{c}{CIFAR-100-LT ($\gamma_l = \gamma_u$)} \\
         \cmidrule(lr){2-16}
         & \multicolumn{3}{c}{Tr-ACC} & \multicolumn{3}{c}{Tr-bACC} & \multicolumn{9}{c}{In-bACC} \\
         \cmidrule(lr){2-4} \cmidrule(lr){5-7} \cmidrule(lr){8-16} 
         Method & All & Old & New & All & Old & New & All & Old & New & KMany & KMed & KFew & UMany & UMed & UFew\\
         \midrule
         $k$-means & 40.1 & 39.6 & 40.6 & 34.2 & 35.0 & 33.4 & 30.3 & 32.9 & 27.6 & 47.9 & 45.0 & 5.8 & 42.9 & 25.7 & 14.2 \\ 
        $\text{ORCA}^{\dagger}$ & 51.2 & 64.9 & 43.9 & 25.0 & 31.5 & 18.6 & 29.5 & 39.1 & 19.9 & 66.1 & 44.6 & 6.6 & 45.8 & 10.8 & 3.1 \\ 
        GCD & \underline{55.0} & 52.1 & \textbf{57.7} & 42.3 & 45.9 & \underline{38.6} & 38.1 & 42.8 & \underline{33.4} & 70.9 & 48.2 & 9.3 & \underline{50.6} & 36.6 & 13.0 \\ 
        $\text{TRSSL}^{\dagger}$ & 41.3 & 73.3 & 25.4 & 33.7 & 46.7 & 20.6 & 37.9 & \underline{53.5} & 22.4 & \underline{80.6} & 51.3 & \underline{28.6} & 35.2 & 25.6 & 6.4 \\ 
        $\text{OpenCon}^{\dagger}$ & 53.5 & \textbf{79.9} & 39.9 & \underline{48.5} & 62.8 & 35.2 & \underline{47.7} & \textbf{62.3} & 33.2 & \textbf{87.3} & \textbf{70.2} & 28.4 & 40.9 & \underline{46.4} & 10.6 \\ 
        PromptCAL & 52.3 & 72.6 & 32.1 & 46.0 & \underline{62.9} & 29.1 & 38.5 & 52.6 & 24.4 & 75.3 & 59.7 & 22.8 & 35.5 & 24.2 & 13.5 \\ 
        SimGCD & 51.7 & 54.3 & 49.2 & 46.5 & 59.8 & 33.2 & 37.4 & 44.1 & 30.8 & 67.1 & 43.9 & 21.3 & 44.6 & 35.2 & 12.6 \\ 
        BaCon & 45.8 & 40.0 & 51.5 & 38.0 & 41.9 & 34.2 & 35.9 & 40.5 & 31.2 & 53.2 & 57.1 & 11.2 & 45.3 & 34.1 & \underline{14.2} \\ 
        \midrule
      \rowcolor{cyan!10}  Ours & \textbf{65.3} & \underline{77.4} & \underline{53.3} & \textbf{53.7} & \textbf{68.4} & \textbf{43.2} & \textbf{48.1} & 52.9 & \textbf{43.2} & 63.2 & \underline{61.1} & \textbf{34.4} & \textbf{55.1} & \textbf{53.3} & \textbf{20.7} \\
         \bottomrule
         
    \end{tabular}
    }
    \label{tab:cifar_mcar}
\end{table*}

\begin{table*}[h]
    \centering
    \caption{Results on CIFAR-100-LT ($\gamma_l \neq \gamma_u$).}
    \resizebox{\textwidth}{!}
    { 
    \begin{tabular}{cccccccccccccccc}
        \toprule
         &  \multicolumn{15}{c}{CIFAR-100-LT ($\gamma_l \neq \gamma_u$)} \\
         \cmidrule(lr){2-16}
         & \multicolumn{3}{c}{Tr-ACC} & \multicolumn{3}{c}{Tr-bACC} & \multicolumn{9}{c}{In-bACC} \\
         \cmidrule(lr){2-4} \cmidrule(lr){5-7} \cmidrule(lr){8-16} 
         Method & All & Old & New & All & Old & New & All & Old & New & KMany & KMed & KFew & UMany & UMed & UFew\\
         \midrule
         $k$-means & 46.0 & 48.4 & 43.6 & 41.8 & 48.4 & 35.2 & 36.9 & 36.9 & 37.0 & 39.2 & 41.9 & 29.6 & 63.0 & 29.3 & 18.7 \\ 
        $\text{ORCA}^{\dagger}$ & 48.8 & 35.5 & 55.5 & 23.8 & 25.5 & 22.2 & 27.2 & 30.5 & 23.8 & 37.0 & 35.1 & 19.4 & 54.2 & 10.9 & 6.3 \\ 
        GCD & 52.8 & 56.8 & 48.9 & 44.3 & 59.7 & 28.9 & 44.6 & 54.0 & 35.1 & 57.9 & 54.9 & 49.3 & \underline{59.2} & 38.2 & 7.9 \\ 
        $\text{TRSSL}^{\dagger}$ & 34.5 & 39.0 & 32.3 & 31.7 & 36.6 & 26.8 & 35.4 & 39.6 & 31.2 & 63.4 & 32.9 & 22.5 & \textbf{62.2} & 20.1 & 11.3 \\ 
        $\text{OpenCon}^{\dagger}$ & 49.6 & 50.7 & 49.0 & 46.3 & 51.1 & \underline{41.5} & 47.4 & 54.3 & \underline{40.4} & \underline{72.3} & 63.0 & 27.6 & 54.9 & \underline{46.8} & \underline{19.5} \\ 
        PromptCal & 56.6 & 76.0 & 37.3 & 54.2 & \textbf{78.0} & 30.4 & 48.1 & \textbf{67.4} & 28.8 & \textbf{80.1} & \textbf{74.3} & 47.8 & 39.5 & 28.2 & 18.7 \\ 
        SimGCD & \underline{65.8} & \textbf{75.2} & \underline{56.4} & \underline{55.2} & 77.0 & 33.4 & \underline{50.3} & 65.3 & 35.4 & 69.4 & 63.9 & \underline{62.6} & 51.0 & 42.4 & 12.8 \\ 
        BaCon & 56.0 & 56.5 & 55.6 & 46.4 & 61.2 & 31.7 & 42.8 & 50.9 & 34.8 & 51.0 & 54.5 & 47.2 & 62.0 & 32.4 & 10.0 \\ 
        \midrule
     \rowcolor{cyan!10}   Ours & \textbf{66.6} & \underline{74.2} & \textbf{59.0} & \textbf{57.3} & 68.7 & \textbf{45.9} & \textbf{53.1} & 64.3 & \textbf{41.8} & 66.0 & \underline{64.1} & \textbf{62.8} & 53.8 & \textbf{49.2} & \textbf{22.4} \\ 
         \bottomrule
         
    \end{tabular}
    }
    \label{tab:cifar_mnar}
\end{table*}

\begin{table*}[h]
    \centering
    \caption{Results on Herbarium19 ($\gamma_l = \gamma_u$). }
    \resizebox{\textwidth}{!}
    { 
    \begin{tabular}{cccccccccccccccc}
        \toprule
         &  \multicolumn{15}{c}{{Herbarium19} ($\gamma_l = \gamma_u$)} \\
         \cmidrule(lr){2-16}
         & \multicolumn{3}{c}{Tr-ACC} & \multicolumn{3}{c}{Tr-bACC} & \multicolumn{9}{c}{In-bACC} \\
         \cmidrule(lr){2-4} \cmidrule(lr){5-7} \cmidrule(lr){8-16} 
         Method & All & Old & New & All & Old & New & All & Old & New & KMany & KMed & KFew & UMany & UMed & UFew\\
         \midrule
         $k$-means & 13.0 & 12.2 & 13.4 & 9.8 & 8.6 & 11.0 & 6.6 & 7.2 & 5.9 & 8.4 & 9.2 & 4.1 & 8.4 & 6.5 & 2.8 \\ 
        $\text{ORCA}^{\dagger}$ & 19.4 & 18.2 & 20.1 & 7.0 & 10.1 & 6.4 & 16.4 & 17.7 & 15.0 & 32.7 & 10.4 & 10.0 & 30.8 & 8.0 & 6.2 \\ 
        GCD & 35.8 & 50.6 & 27.8 & 33.4 & 42.3 & 24.5 & 25.5 & 25.1 & 25.9 & 36.1 & 24.0 & 15.2 & 36.0 & 29.3 & 12.4 \\ 
        $\text{TRSSL}^{\dagger}$ & 40.2 & \textbf{67.2} & 16.4 & 32.0 & \textbf{54.0} & 10.0 & 33.3 & 33.4 & 33.3 & \textbf{56.3} & 31.9 & 12.0 & 49.7 & 36.0 & 14.2 \\ 
        $\text{OpenCon}^{\dagger}$ & 28.6 & 46.2 & 19.2 & 20.9 & 31.5 & 10.4 & 29.7 & 27.8 & 31.7 & 39.8 & 23.4 & 20.2 & 50.8 & 39.5 & \underline{29.1} \\ 
        PromptCAL & 34.1 & 49.7 & 25.7 & \underline{34.4} & 44.3 & 24.5 & 32.0 & 33.1 & 30.9 & 42.6 & 33.1 & \underline{23.6} & 41.6 & 30.7 & 20.4 \\ 
        SimGCD & \underline{43.4} & \underline{57.7} & \underline{35.8} & 33.9 & \underline{45.8} & 22.1 & \underline{42.3} & \underline{40.1} & \underline{44.6} & 55.8 & \underline{42.8} & 21.7 & \underline{60.6} & \underline{49.4} & 23.8 \\ 
        BaCon & 29.8 & 29.2 & 30.1 & 28.7 & 28.3 & \underline{29.2} & 27.1 & 27.1 & 27.2 & 45.3 & 23.4 & 12.5 & 38.1 & 26.9 & 16.6 \\ 
        \midrule
       \rowcolor{cyan!10} Ours & \textbf{47.7} & 48.7 & \textbf{46.8} & \textbf{38.9} & 39.6 & \textbf{38.1} & \textbf{45.4} & \textbf{43.0} & \textbf{47.8} & \underline{56.1} & \textbf{44.8} & \textbf{28.0} & \textbf{64.1} & \textbf{49.7} & \textbf{29.5} \\
         \bottomrule
    \end{tabular}
    }
    \label{tab:herb_mcar}
\end{table*}

\clearpage

\begin{table*}[t]
    \centering
    \caption{Results on Herbarium19 ($\gamma_l \neq \gamma_u$).}
    \resizebox{\textwidth}{!}
    { 
    \begin{tabular}{cccccccccccccccc}
        \toprule
         &  \multicolumn{15}{c}{{Herbarium19} ($\gamma_l = \gamma_u$)} \\
         \cmidrule(lr){2-16}
         & \multicolumn{3}{c}{Tr-ACC} & \multicolumn{3}{c}{Tr-bACC} & \multicolumn{9}{c}{In-bACC} \\
         \cmidrule(lr){2-4} \cmidrule(lr){5-7} \cmidrule(lr){8-16} 
         Method & All & Old & New & All & Old & New & All & Old & New & KMany & KMed & KFew & UMany & UMed & UFew\\
         \midrule
         GCD & 27.3 & 32.8 & 24.3 & 28.9 & 34.2 & 23.6 & 18.2 & 23.0 & 13.4 & 25.4 & 29.1 & 14.3 & 19.2 & 11.8 & 9.1 \\ 
        SimGCD & \underline{34.9} & \underline{40.9} & \underline{31.7} & 31.2 & \underline{38.2} & 24.3 & \underline{26.5} & \underline{32.5} & \underline{20.5} & 39.2 & \underline{38.5} & \underline{19.6} & \underline{32.1} & \underline{16.3} & \underline{13.1} \\ 
        BaCon & 32.4 & 35.6 & 30.6 & \underline{31.6} & 35.3 & \underline{27.9} & 21.5 & 26.7 & 16.3 & \underline{39.6} & 25.6 & 14.9 & 28.2 & 14.3 & 6.4 \\ 
        \midrule
      \rowcolor{cyan!10}  Ours & \textbf{46.9} & \textbf{58.4} & \textbf{40.8} & \textbf{37.0} & \textbf{48.9} & \underline{25.1} & \textbf{31.4} & \textbf{41.5} & \textbf{21.3} & \textbf{57.2} & \textbf{39.7} & \textbf{27.4} & \textbf{32.9} & \textbf{16.5} & \textbf{14.7} \\
         \bottomrule
         
    \end{tabular}
    }
    \label{tab:herb_mnar}
\end{table*}

\section{Results on ImageNet-100-LT}
While ImageNet-100 has been often used for OWSSL in literature,
we argue that ImageNet-100 might not be appropriate for the conventional OWSSL settings built on top of ImageNet-1K~\cite{5206848} pre-trained backbone, \eg, DINO-ViT~\cite{caron2021emerging}, as it already observed data from novel classes during pretraining.
In other words, the performance could be boosted by preserving the pre-trained knowledge rather than learning to discover novel classes and classify all classes.
Nevertheless, below we report the performance on ImageNet-100 with the data split from BaCon~\cite{bai2023effectiveness}.
In~\Crefrange{tab:in100_mcar}{tab:in100_mnar}, our proposed method achieves significantly better performance on novel and tail classes, surpassing the baseline performance.
Notably, the classic baseline GCD~\cite{vaze2022generalized} often shows the best performance (mostly in transductive inference), implying that it preserves the pre-trained knowledge better.
\begin{table*}[h]
    \centering
    \caption{Results on ImageNet-100-LT ($\gamma_l = \gamma_u$). }
    \resizebox{\textwidth}{!}
    { 
    \begin{tabular}{cccccccccccccccc}
        \toprule
         &  \multicolumn{15}{c}{ImageNet-100-LT ($\gamma_l = \gamma_u$)} \\
         \cmidrule(lr){2-16}
         & \multicolumn{3}{c}{Tr-ACC} & \multicolumn{3}{c}{Tr-bACC} & \multicolumn{9}{c}{In-bACC} \\
         \cmidrule(lr){2-4} \cmidrule(lr){5-7} \cmidrule(lr){8-16} 
         Method & All & Old & New & All & Old & New & All & Old & New & KMany & KMed & KFew & UMany & UMed & UFew\\
         \midrule
          GCD  & \underline{63.8} & \underline{69.5}& \textbf{60.7} & \textbf{63.5}& \underline{69.4}& \textbf{57.6}& \underline{59.2} & \underline{67.1} & \underline{51.3} & 81.1 & 77.1 & \underline{41.8} & \textbf{76.0} & \underline{56.8} & 20.5 \\
          SimGCD  & 55.3 & 62.3 & 51.6 & 55.5 & 63.9 & 47.0 &  54.0 & 63.2 & 44.8 & 64.0 & 80.2 & \textbf{43.4} & 64.8 & 48.2 & 20.9 \\
          BaCon & 60.7 & 68.8 & 56.4 & 58.6 & 66.9 & 50.4 & 54.8 & 65.8 & 43.7 & \underline{81.5} & \underline{81.0} & 33.0 & 67.8 & 41.3 & \underline{22.4} \\
         \midrule
       \rowcolor{cyan!10}  Ours  & \textbf{65.6} & \textbf{82.5} & \underline{56.6} &\underline{63.0} & \textbf{71.2} & \underline{54.7} & \textbf{61.4} & \textbf{69.6} & \textbf{53.3} & \textbf{85.2} & \textbf{81.2} & 40.9 & \underline{70.0} & \textbf{64.6} & \textbf{23.6} \\ 
         \bottomrule
         
    \end{tabular}
    }
    \label{tab:in100_mcar}
\end{table*}

\begin{table*}[h]
    \centering
    \caption{Results on ImageNet-100-LT ($\gamma_l \neq \gamma_u$). }
    \resizebox{\textwidth}{!}
    { 
    \begin{tabular}{cccccccccccccccc}
        \toprule
         &  \multicolumn{15}{c}{ImageNet-100-LT ($\gamma_l \neq \gamma_u$)} \\
         \cmidrule(lr){2-16}
         & \multicolumn{3}{c}{Tr-ACC} & \multicolumn{3}{c}{Tr-bACC} & \multicolumn{9}{c}{In-bACC} \\
         \cmidrule(lr){2-4} \cmidrule(lr){5-7} \cmidrule(lr){8-16} 
         Method & All & Old & New & All & Old & New & All & Old & New & KMany & KMed & KFew & UMany & UMed & UFew\\
         \midrule
          GCD & \textbf{66.2} & \textbf{75.9} & \textbf{61.1} & 62.1 & 67.2 & 57.0 & 60.6 & 66.2 & 55.0 & \textbf{87.1} & 66.8 & \textbf{44.8} & \textbf{73.4} & \textbf{62.2} & 28.5 \\ 
          SimGCD & 60.4 & \underline{75.4} & 52.4 & 56.5 & 66.9 & 46.2 & 56.0 & 66.6 & 45.4 & 75.6 & \underline{84.2} & 37.6 & 69.2 & 53.1 & 12.8 \\ 
          BaCon & 60.7 & 68.8 & 56.4 & 58.6 & 66.9 & 50.4 & 54.7 & 65.8 & 43.7 & \underline{81.5} & 81.0 & 33.0 & 67.8 & 41.3 & 22.4 \\ \midrule
        \rowcolor{cyan!10}  Ours & \underline{63.2} & 73.9 & \underline{57.4} & \textbf{63.3} & \textbf{68.9} & \textbf{57.6} & \textbf{62.9} & \textbf{68.5} & \textbf{57.2} & 77.6 & \textbf{86.4} & \underline{41.5} & \underline{71.4} & \underline{55.3} & \textbf{45.2} \\
         \bottomrule
         
    \end{tabular}
    }
    \label{tab:in100_mnar}
\end{table*}

\newpage

\section{Visualizations}
\begin{figure*}[t!]
    \centering
    \begin{subfigure}[b]{0.23\textwidth}
        \includegraphics[width=\textwidth]{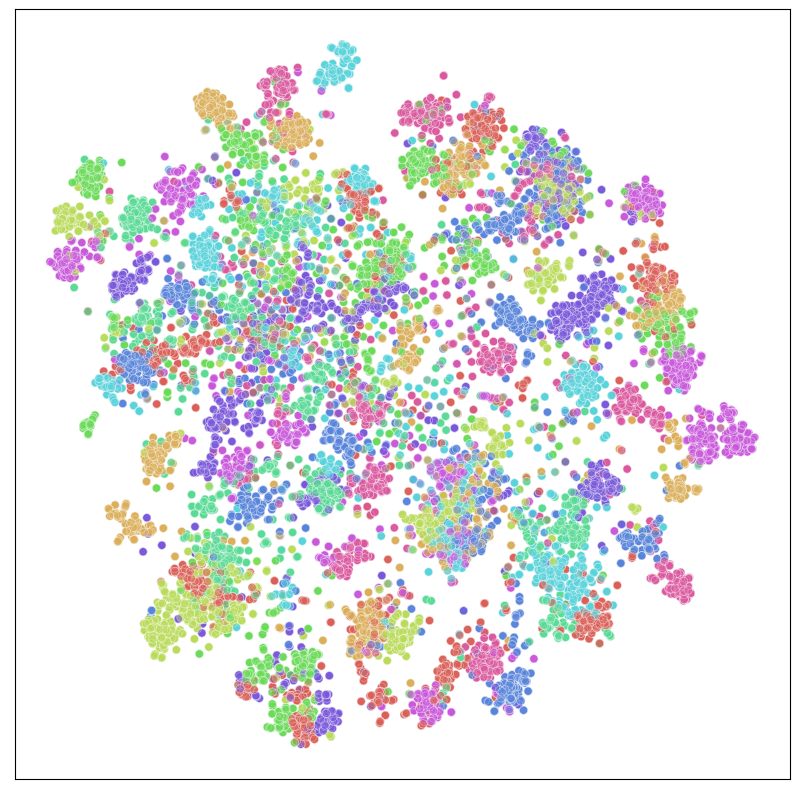}
        \caption{DINO}
        \label{fig:kmeans}
    \end{subfigure}
    \begin{subfigure}[b]{0.23\textwidth}
        \includegraphics[width=\textwidth]{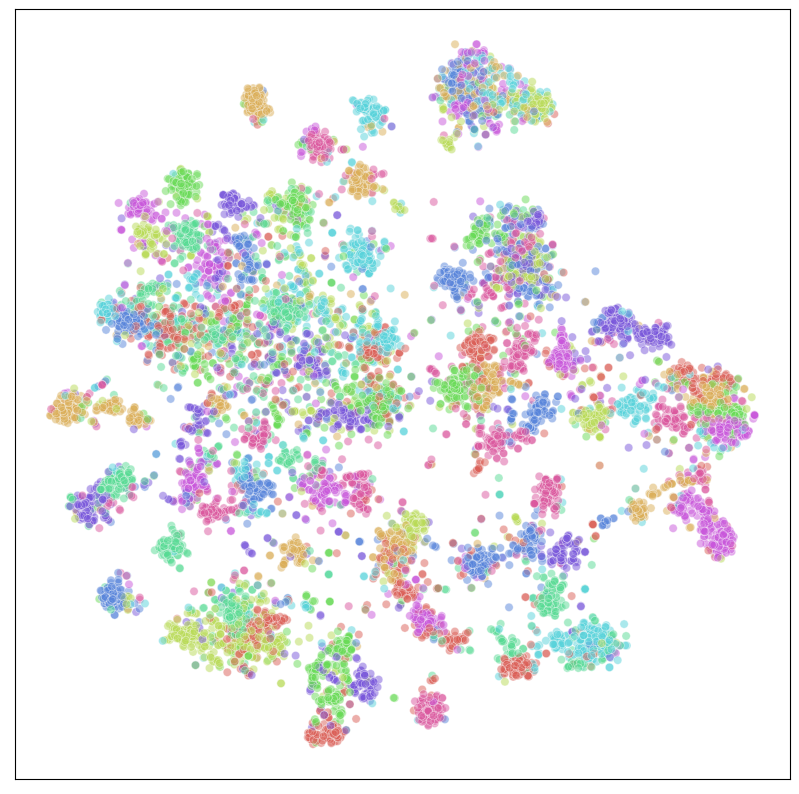}
        \caption{GCD}
        \label{fig:gcd}
    \end{subfigure}
    \begin{subfigure}[b]{0.23\textwidth}
        \includegraphics[width=\textwidth]{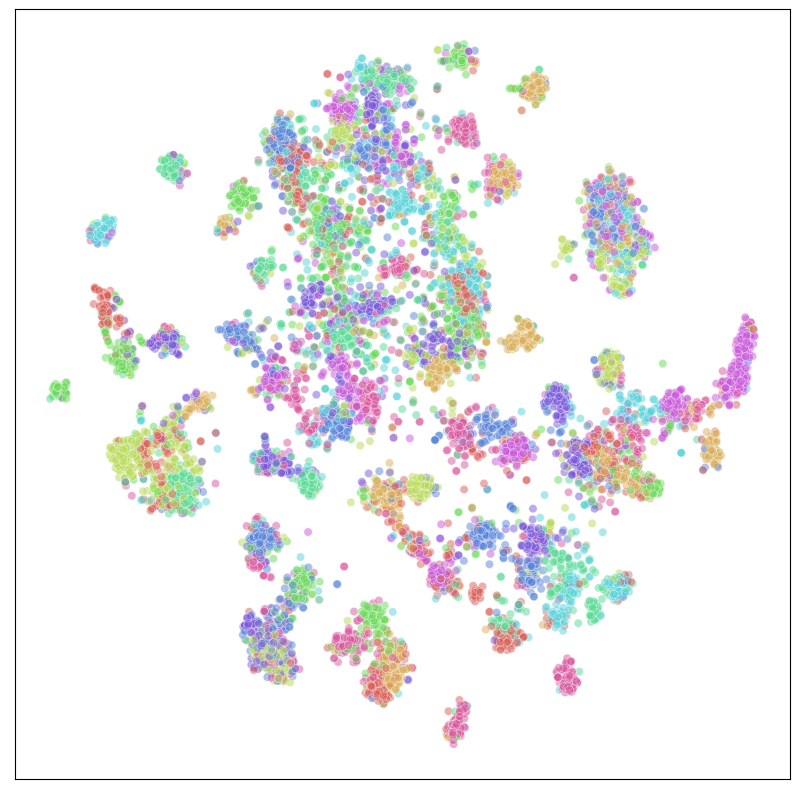}
        \caption{SimGCD}
        \label{fig:simgcd}
    \end{subfigure}
     \begin{subfigure}[b]{0.23\textwidth}
        \includegraphics[width=\textwidth]{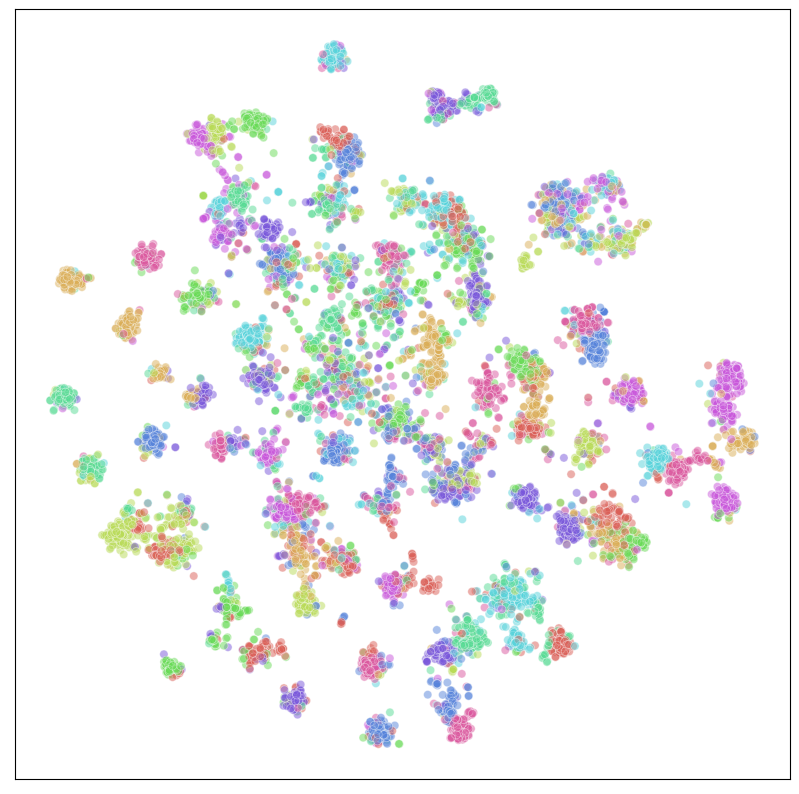}
        \caption{DTS (Ours)}
        \label{fig:ours}
    \end{subfigure}
    \caption{t-SNE visualization on the test set of CIFAR-100-LT.}
    \label{fig:tsne}
\end{figure*}

To inspect the learned semantic discriminativeness of
the proposed DTS on the long-tailed dataset, we visualize embeddings by t-SNE~\cite{van2008visualizing} algorithm, trained on CIFAR-100-LT with distribution match.
We show the feature embedding of pretrained DINO~\cite{caron2021emerging}, GCD~\cite{vaze2022generalized}, SimGCD~\cite{wen2022simple}, and DTS (Ours), in~\figref{fig:tsne}.
Compared to other models, the model trained with our DTS learns less ambiguous features which exhibits a larger margin between different classes, with more compact clusters. 
This indicates that our method is more effective in learning a discriminative semantic structure,
even under long-tailed datasets.

\section{Conclusion and Limitations}

In this paper, we formulate the practical ROWSSL setting, which considers the long-tailed distribution and the class prior distribution mismatch between labeled and unlabeled data for training, and inductive and transductive inferences for evaluation.
To tackle ROWSSL, we introduce
a novel method called Density-based Temperature scaling and Soft pseudo-labeling (DTS), 
which learns class-balanced representations
and mitigates the classification bias based on local densities.
Nevertheless, we acknowledge several limitations inherent in DTS and existing methods.
First, the labeled and unlabeled data are sampled from the same dataset,
which might not reflect domain shifts in real-world scenarios. 
Second, estimating the number of novel classes with off-the-shelf methods can result in inaccurate prediction due to the imbalanced class prior distribution.
We believe that ROWSSL will establish a robust foundation for future research
and contribute to the development of more reliable methods for practical applications of OWSSL.

\section{Negative Societal Impact}
While our work itself is not inherently harmful to society, there is a risk that it could be misused by those with malicious intent. 
For example, the proposed method could be used to unfairly single out and target certain groups, such as minorities. 
Consequently, we urge that this work must be utilized within ethical and legal boundaries.

\clearpage

%% file: main.bbl
\begin{thebibliography}{56}
\providecommand{\natexlab}[1]{#1}
\providecommand{\url}[1]{\texttt{#1}}
\expandafter\ifx\csname urlstyle\endcsname\relax
  \providecommand{\doi}[1]{doi: #1}\else
  \providecommand{\doi}{doi: \begingroup \urlstyle{rm}\Url}\fi

\bibitem[Arazo et~al.(2020)Arazo, Ortego, Albert, O’Connor, and McGuinness]{arazo2020pseudo}
Eric Arazo, Diego Ortego, Paul Albert, Noel~E O’Connor, and Kevin McGuinness.
\newblock Pseudo-labeling and confirmation bias in deep semi-supervised learning.
\newblock \emph{{\normalfont In} IJCNN}, 2020.

\bibitem[Assran et~al.(2023)Assran, Balestriero, Duval, Bordes, Misra, Bojanowski, Vincent, Rabbat, and Ballas]{assran2022hidden}
Mahmoud Assran, Randall Balestriero, Quentin Duval, Florian Bordes, Ishan Misra, Piotr Bojanowski, Pascal Vincent, Michael Rabbat, and Nicolas Ballas.
\newblock The hidden uniform cluster prior in self-supervised learning.
\newblock \emph{{\normalfont In} ICLR}, 2023.

\bibitem[Bai et~al.(2023{\natexlab{a}})Bai, Liu, Wang, Chen, Mu, Li, Zhou, Feng, Wu, and Hu]{bai2023towards}
Jianhong Bai, Zuozhu Liu, Hualiang Wang, Ruizhe Chen, Lianrui Mu, Xiaomeng Li, Joey~Tianyi Zhou, Yang Feng, Jian Wu, and Haoji Hu.
\newblock Towards distribution-agnostic generalized category discovery.
\newblock \emph{arXiv preprint arXiv:2310.01376}, 2023{\natexlab{a}}.

\bibitem[Bai et~al.(2023{\natexlab{b}})Bai, Liu, Wang, Hao, Feng, Chu, and Hu]{bai2023effectiveness}
Jianhong Bai, Zuozhu Liu, Hualiang Wang, Jin Hao, Yang Feng, Huanpeng Chu, and Haoji Hu.
\newblock On the effectiveness of out-of-distribution data in self-supervised long-tail learning.
\newblock \emph{{\normalfont In} ICLR}, 2023{\natexlab{b}}.

\bibitem[Cao et~al.(2019)Cao, Wei, Gaidon, Arechiga, and Ma]{cao2019learning}
Kaidi Cao, Colin Wei, Adrien Gaidon, Nikos Arechiga, and Tengyu Ma.
\newblock Learning imbalanced datasets with label-distribution-aware margin loss.
\newblock \emph{{\normalfont In} NeurIPS}, 2019.

\bibitem[Cao et~al.(2021)Cao, Brbic, and Leskovec]{cao2021open}
Kaidi Cao, Maria Brbic, and Jure Leskovec.
\newblock Open-world semi-supervised learning.
\newblock \emph{{\normalfont In} ICLR}, 2021.

\bibitem[Caron et~al.(2021)Caron, Touvron, Misra, J{\'e}gou, Mairal, Bojanowski, and Joulin]{caron2021emerging}
Mathilde Caron, Hugo Touvron, Ishan Misra, Herv{\'e} J{\'e}gou, Julien Mairal, Piotr Bojanowski, and Armand Joulin.
\newblock Emerging properties in self-supervised vision transformers.
\newblock \emph{{\normalfont In} ICCV}, 2021.

\bibitem[Chawla et~al.(2002)Chawla, Bowyer, Hall, and Kegelmeyer]{chawla2002smote}
N.~V. Chawla, K.~W. Bowyer, L.~O. Hall, and W.~P. Kegelmeyer.
\newblock {SMOTE}: Synthetic minority over-sampling technique.
\newblock \emph{Journal of Artificial Intelligence Research}, 2002.

\bibitem[Chuyu et~al.(2023)Chuyu, Ruijie, and Xuming]{chuyu2023novel}
Zhang Chuyu, Xu Ruijie, and He Xuming.
\newblock Novel class discovery for long-tailed recognition.
\newblock \emph{{\normalfont In} TMLR}, 2023.

\bibitem[Cui et~al.(2019)Cui, Jia, Lin, Song, and Belongie]{cui2019classbalanced}
Yin Cui, Menglin Jia, Tsung-Yi Lin, Yang Song, and Serge Belongie.
\newblock Class-balanced loss based on effective number of samples.
\newblock \emph{{\normalfont In} CVPR}, 2019.

\bibitem[Dai et~al.(2022)Dai, Cai, and Chen]{9945500}
Zhigang Dai, Bolun Cai, and Junying Chen.
\newblock Unimoco: Unsupervised, semi-supervised and fully-supervised visual representation learning.
\newblock \emph{2022 IEEE International Conference on Systems, Man, and Cybernetics (SMC)}, 2022.

\bibitem[Deng et~al.(2009)Deng, Dong, Socher, Li, Li, and Fei-Fei]{5206848}
Jia Deng, Wei Dong, Richard Socher, Li-Jia Li, Kai Li, and Li Fei-Fei.
\newblock Imagenet: A large-scale hierarchical image database.
\newblock \emph{{\normalfont In} CVPR}, 2009.

\bibitem[Dosovitskiy et~al.(2021)Dosovitskiy, Beyer, Kolesnikov, Weissenborn, Zhai, Unterthiner, Dehghani, Minderer, Heigold, Gelly, et~al.]{dosovitskiy2020image}
Alexey Dosovitskiy, Lucas Beyer, Alexander Kolesnikov, Dirk Weissenborn, Xiaohua Zhai, Thomas Unterthiner, Mostafa Dehghani, Matthias Minderer, Georg Heigold, Sylvain Gelly, et~al.
\newblock An image is worth 16x16 words: Transformers for image recognition at scale.
\newblock \emph{{\normalfont In} ICLR}, 2021.

\bibitem[Duan et~al.(2023)Duan, Zhao, Qi, Zhou, Wang, and Shi]{duan2023towards}
Yue Duan, Zhen Zhao, Lei Qi, Luping Zhou, Lei Wang, and Yinghuan Shi.
\newblock Towards semi-supervised learning with non-random missing labels.
\newblock \emph{{\normalfont In} ICCV}, 2023.

\bibitem[Dudani(1976)]{Dudani1976wknn}
Sahibsingh~A. Dudani.
\newblock The distance-weighted k-nearest-neighbor rule.
\newblock \emph{IEEE Transactions on Systems, Man, and Cybernetics}, 1976.

\bibitem[Fei et~al.(2022)Fei, Zhao, Yang, and Zhao]{fei2022xcon}
Yixin Fei, Zhongkai Zhao, Siwei Yang, and Bingchen Zhao.
\newblock Xcon: Learning with experts for fine-grained category discovery.
\newblock \emph{{\normalfont In} BMVC}, 2022.

\bibitem[Fini et~al.(2023)Fini, Astolfi, Alahari, Alameda-Pineda, Mairal, Nabi, and Ricci]{fini2023semi}
Enrico Fini, Pietro Astolfi, Karteek Alahari, Xavier Alameda-Pineda, Julien Mairal, Moin Nabi, and Elisa Ricci.
\newblock Semi-supervised learning made simple with self-supervised clustering.
\newblock \emph{{\normalfont In} CVPR}, 2023.

\bibitem[Guo et~al.(2020)Guo, Zhang, Jiang, Li, and Zhou]{guo2020safe}
Lan-Zhe Guo, Zhen-Yu Zhang, Yuan Jiang, Yu-Feng Li, and Zhi-Hua Zhou.
\newblock Safe deep semi-supervised learning for unseen-class unlabeled data.
\newblock \emph{{\normalfont In} ICML}, 2020.

\bibitem[Guo et~al.(2022)Guo, Zhang, Wu, Shao, and Li]{guo2022robust}
Lan-Zhe Guo, Yi-Ge Zhang, Zhi-Fan Wu, Jie-Jing Shao, and Yu-Feng Li.
\newblock Robust semi-supervised learning when not all classes have labels.
\newblock \emph{{\normalfont In} NeurIPS}, 2022.

\bibitem[Han et~al.(2019)Han, Vedaldi, and Zisserman]{han2019learning}
Kai Han, Andrea Vedaldi, and Andrew Zisserman.
\newblock Learning to discover novel visual categories via deep transfer clustering.
\newblock \emph{{\normalfont In} ICCV}, 2019.

\bibitem[Hao et~al.(2023)Hao, Han, and Wong]{hao2023cipr}
Shaozhe Hao, Kai Han, and Kwan-Yee~K Wong.
\newblock Cipr: An efficient framework with cross-instance positive relations for generalized category discovery.
\newblock \emph{arXiv preprint arXiv:2304.06928}, 2023.

\bibitem[He et~al.(2020)He, Fan, Wu, Xie, and Girshick]{he2020momentum}
Kaiming He, Haoqi Fan, Yuxin Wu, Saining Xie, and Ross Girshick.
\newblock Momentum contrast for unsupervised visual representation learning.
\newblock \emph{{\normalfont In} CVPR}, 2020.

\bibitem[Hsu and Kira(2016)]{hsu2015neural}
Yen-Chang Hsu and Zsolt Kira.
\newblock Neural network-based clustering using pairwise constraints.
\newblock \emph{{\normalfont In} ICLR Workshop}, 2016.

\bibitem[Hsu et~al.(2019)Hsu, Lv, Schlosser, Odom, and Kira]{hsu2019multi}
Yen-Chang Hsu, Zhaoyang Lv, Joel Schlosser, Phillip Odom, and Zsolt Kira.
\newblock Multi-class classification without multi-class labels.
\newblock \emph{{\normalfont In} ICLR}, 2019.

\bibitem[Hu et~al.(2022)Hu, Niu, Miao, Hua, and Zhang]{hu2022non}
Xinting Hu, Yulei Niu, Chunyan Miao, Xian-Sheng Hua, and Hanwang Zhang.
\newblock On non-random missing labels in semi-supervised learning.
\newblock \emph{{\normalfont In} ICLR}, 2022.

\bibitem[Jiang et~al.(2021)Jiang, Chen, Mortazavi, and Wang]{jiang2021self}
Ziyu Jiang, Tianlong Chen, Bobak~J Mortazavi, and Zhangyang Wang.
\newblock Self-damaging contrastive learning.
\newblock \emph{{\normalfont In} ICML}, 2021.

\bibitem[Khosla et~al.(2020)Khosla, Teterwak, Wang, Sarna, Tian, Isola, Maschinot, Liu, and Krishnan]{khosla2020supervised}
Prannay Khosla, Piotr Teterwak, Chen Wang, Aaron Sarna, Yonglong Tian, Phillip Isola, Aaron Maschinot, Ce Liu, and Dilip Krishnan.
\newblock Supervised contrastive learning.
\newblock \emph{{\normalfont In} NeurIPS}, 2020.

\bibitem[Kim et~al.(2020)Kim, Hur, Park, Yang, Hwang, and Shin]{kim2020distribution}
Jaehyung Kim, Youngbum Hur, Sejun Park, Eunho Yang, Sung~Ju Hwang, and Jinwoo Shin.
\newblock Distribution aligning refinery of pseudo-label for imbalanced semi-supervised learning.
\newblock \emph{{\normalfont In} NeurIPS}, 2020.

\bibitem[Kuhn(1955)]{kuhn1955hungarian}
Harold~W Kuhn.
\newblock The hungarian method for the assignment problem.
\newblock \emph{Naval research logistics quarterly}, 1955.

\bibitem[Kukleva et~al.(2023)Kukleva, B{\"o}hle, Schiele, Kuehne, and Rupprecht]{kukleva2023temperature}
Anna Kukleva, Moritz B{\"o}hle, Bernt Schiele, Hilde Kuehne, and Christian Rupprecht.
\newblock Temperature schedules for self-supervised contrastive methods on long-tail data.
\newblock \emph{{\normalfont In} ICLR}, 2023.

\bibitem[Liang et~al.(2012)Liang, Bai, Dang, and Cao]{liang2012k}
Jiye Liang, Liang Bai, Chuangyin Dang, and Fuyuan Cao.
\newblock The $ k $-means-type algorithms versus imbalanced data distributions.
\newblock \emph{IEEE Transactions on Fuzzy Systems}, 2012.

\bibitem[Liu et~al.(2022)Liu, HaoChen, Gaidon, and Ma]{liu2022selfsupervised}
Hong Liu, Jeff~Z. HaoChen, Adrien Gaidon, and Tengyu Ma.
\newblock Self-supervised learning is more robust to dataset imbalance.
\newblock \emph{{\normalfont In} ICLR}, 2022.

\bibitem[Liu et~al.(2023)Liu, Wang, Zhang, Fan, Yang, and Shao]{liu2023open}
Jiaming Liu, Yangqiming Wang, Tongze Zhang, Yulu Fan, Qinli Yang, and Junming Shao.
\newblock Open-world semi-supervised novel class discovery.
\newblock \emph{{\normalfont In} IJCAI}, 2023.

\bibitem[MacQueen et~al.(1967)]{macqueen1967some}
James MacQueen et~al.
\newblock Some methods for classification and analysis of multivariate observations.
\newblock \emph{Proceedings of the fifth Berkeley symposium on mathematical statistics and probability}, 1967.

\bibitem[Menon et~al.(2021)Menon, Jayasumana, Rawat, Jain, Veit, and Kumar]{menon2020long}
Aditya~Krishna Menon, Sadeep Jayasumana, Ankit~Singh Rawat, Himanshu Jain, Andreas Veit, and Sanjiv Kumar.
\newblock Long-tail learning via logit adjustment.
\newblock \emph{{\normalfont In} ICLR}, 2021.

\bibitem[Oliver et~al.(2018)Oliver, Odena, Raffel, Cubuk, and Goodfellow]{oliver2018realistic}
Avital Oliver, Augustus Odena, Colin~A Raffel, Ekin~Dogus Cubuk, and Ian Goodfellow.
\newblock Realistic evaluation of deep semi-supervised learning algorithms.
\newblock \emph{{\normalfont In} NeurIPS}, 2018.

\bibitem[Paszke et~al.(2019)Paszke, Gross, Massa, Lerer, Bradbury, Chanan, Killeen, Lin, Gimelshein, Antiga, et~al.]{paszke2019pytorch}
Adam Paszke, Sam Gross, Francisco Massa, Adam Lerer, James Bradbury, Gregory Chanan, Trevor Killeen, Zeming Lin, Natalia Gimelshein, Luca Antiga, et~al.
\newblock Pytorch: An imperative style, high-performance deep learning library.
\newblock \emph{{\normalfont In} NeurIPS}, 2019.

\bibitem[Pu et~al.(2023)Pu, Zhong, and Sebe]{pu2023dynamic}
Nan Pu, Zhun Zhong, and Nicu Sebe.
\newblock Dynamic conceptional contrastive learning for generalized category discovery.
\newblock \emph{{\normalfont In} CVPR}, 2023.

\bibitem[Qian et~al.(2019)Qian, Shang, Sun, Hu, Li, and Jin]{qian2019softtriple}
Qi Qian, Lei Shang, Baigui Sun, Juhua Hu, Hao Li, and Rong Jin.
\newblock Softtriple loss: Deep metric learning without triplet sampling.
\newblock \emph{{\normalfont In} ICCV}, 2019.

\bibitem[Rizve et~al.(2022{\natexlab{a}})Rizve, Kardan, Khan, Shahbaz~Khan, and Shah]{rizve2022openldn}
Mamshad~Nayeem Rizve, Navid Kardan, Salman Khan, Fahad Shahbaz~Khan, and Mubarak Shah.
\newblock Openldn: Learning to discover novel classes for open-world semi-supervised learning.
\newblock \emph{{\normalfont In} ECCV}, 2022{\natexlab{a}}.

\bibitem[Rizve et~al.(2022{\natexlab{b}})Rizve, Kardan, and Shah]{rizve2022realistic}
Mamshad~Nayeem Rizve, Navid Kardan, and Mubarak Shah.
\newblock Towards realistic semi-supervised learning.
\newblock \emph{{\normalfont In} ECCV}, 2022{\natexlab{b}}.

\bibitem[Sun and Li(2023)]{sun2022opencon}
Yiyou Sun and Yixuan Li.
\newblock Opencon: Open-world contrastive learning.
\newblock \emph{{\normalfont In} TMLR}, 2023.

\bibitem[Van~der Maaten and Hinton(2008)]{van2008visualizing}
Laurens Van~der Maaten and Geoffrey Hinton.
\newblock Visualizing data using t-sne.
\newblock \emph{{\normalfont In} JMLR}, 2008.

\bibitem[Vaze et~al.(2022)Vaze, Han, Vedaldi, and Zisserman]{vaze2022generalized}
Sagar Vaze, Kai Han, Andrea Vedaldi, and Andrew Zisserman.
\newblock Generalized category discovery.
\newblock \emph{{\normalfont In} CVPR}, 2022.

\bibitem[Wang and Liu(2021)]{wang2021understanding}
Feng Wang and Huaping Liu.
\newblock Understanding the behaviour of contrastive loss.
\newblock \emph{{\normalfont In} CVPR}, 2021.

\bibitem[Wang et~al.(2022)Wang, Zhang, Zhu, Zheng, Li, Smola, and Wang]{wang2022partial}
Haotao Wang, Aston Zhang, Yi Zhu, Shuai Zheng, Mu Li, Alex~J Smola, and Zhangyang Wang.
\newblock Partial and asymmetric contrastive learning for out-of-distribution detection in long-tailed recognition.
\newblock \emph{{\normalfont In} ICML}, 2022.

\bibitem[Wang et~al.(2024)Wang, Vaze, and Han]{wang2024sptnet}
Hongjun Wang, Sagar Vaze, and Kai Han.
\newblock Sptnet: An efficient alternative framework for generalized category discovery with spatial prompt tuning.
\newblock \emph{{\normalfont In} ICLR}, 2024.

\bibitem[Wang and Isola(2020)]{wang2020understanding}
Tongzhou Wang and Phillip Isola.
\newblock Understanding contrastive representation learning through alignment and uniformity on the hypersphere.
\newblock \emph{{\normalfont In} ICML}, 2020.

\bibitem[Wei et~al.(2021)Wei, Sohn, Mellina, Yuille, and Yang]{wei2021crest}
Chen Wei, Kihyuk Sohn, Clayton Mellina, Alan Yuille, and Fan Yang.
\newblock Crest: A class-rebalancing self-training framework for imbalanced semi-supervised learning.
\newblock \emph{{\normalfont In} CVPR}, 2021.

\bibitem[Wei and Gan(2023)]{wei2023towards}
Tong Wei and Kai Gan.
\newblock Towards realistic long-tailed semi-supervised learning: Consistency is all you need.
\newblock \emph{{\normalfont In} CVPR}, 2023.

\bibitem[Wen et~al.(2023)Wen, Zhao, and Qi]{wen2022simple}
Xin Wen, Bingchen Zhao, and Xiaojuan Qi.
\newblock A simple parametric classification baseline for generalized category discovery.
\newblock \emph{{\normalfont In} ICCV}, 2023.

\bibitem[Wu et~al.(2009)Wu, Xiong, and Chen]{wu2009adapting}
Junjie Wu, Hui Xiong, and Jian Chen.
\newblock Adapting the right measures for k-means clustering.
\newblock \emph{{\normalfont In} KDD}, 2009.

\bibitem[Yang et~al.(2023)Yang, Wang, Deng, and Zhang]{yang2023bootstrap}
Muli Yang, Liancheng Wang, Cheng Deng, and Hanwang Zhang.
\newblock Bootstrap your own prior: Towards distribution-agnostic novel class discovery.
\newblock \emph{{\normalfont In} CVPR}, 2023.

\bibitem[Zhang et~al.(2023)Zhang, Khan, Shen, Naseer, Chen, and Khan]{zhang2023promptcal}
Sheng Zhang, Salman Khan, Zhiqiang Shen, Muzammal Naseer, Guangyi Chen, and Fahad~Shahbaz Khan.
\newblock Promptcal: Contrastive affinity learning via auxiliary prompts for generalized novel category discovery.
\newblock \emph{{\normalfont In} CVPR}, 2023.

\bibitem[Zhao et~al.(2023)Zhao, Wen, and Han]{zhao2023learning}
Bingchen Zhao, Xin Wen, and Kai Han.
\newblock Learning semi-supervised gaussian mixture models for generalized category discovery.
\newblock \emph{{\normalfont In} ICCV}, 2023.

\bibitem[Zhou et~al.(2022)Zhou, Yao, Wang, Han, and Zhang]{zhou2022contrastive}
Zhihan Zhou, Jiangchao Yao, Yan-Feng Wang, Bo Han, and Ya Zhang.
\newblock Contrastive learning with boosted memorization.
\newblock \emph{{\normalfont In} ICML}, 2022.

\end{thebibliography}
